\documentclass[10pt,twocolumn,letterpaper]{article}

\usepackage[pagenumbers]{iccv} % To force page numbers, e.g. for an arXiv version

\usepackage{pifont}
\usepackage{graphicx}
\usepackage{amssymb}
\usepackage{booktabs}
\usepackage{currfile}

\usepackage{multirow}
\usepackage{colortbl}
\usepackage{gensymb}
\usepackage{algorithm}
\usepackage{algpseudocode}
\usepackage{comment} % ADDED
\usepackage[dvipsnames]{xcolor}
\usepackage[accsupp]{axessibility}

\definecolor{iccvblue}{rgb}{0.21,0.49,0.74}
\usepackage[pagebackref,breaklinks,colorlinks,allcolors=iccvblue]{hyperref}

\definecolor{tabfirst}{rgb}{0.196, 0.607, 0.317} % solid green BOLT
\definecolor{tabsecond}{rgb}{0.454, 0.768, 0.462} % half solid green
\definecolor{tabthird}{rgb}{0.855, 0.943, 0.833} % QUITE LIGHT

\iftrue  % \iffalse
\newcommand{\xcyan}[1]{{\bf \color{purple}[#1]}}
\newcommand{\akshay}[1]{{\bf \color{cyan}[#1]}}
\else
\newcommand{\xcyan}[1]{{}}
\newcommand{\akshay}[1]{{}}
\fi

\iftrue  % \iffalse
\newcommand{\casser}[1]{{\bf \color{magenta}[#1]}}
\else
\newcommand{\casser}[1]{{}}
\fi

\iftrue
\newcommand{\cutsectionup}{\vspace*{-8pt}} % -10
\newcommand{\cutsectiondown}{\vspace*{-4pt}}
\newcommand{\cutsubsectionup}{\vspace*{-4pt}}
\newcommand{\cutsubsectiondown}{\vspace*{-4pt}}

\newcommand{\cutparagraphup}{\vspace*{-7pt}} % -10

\newcommand{\cutcaptionup}{\vspace*{-8pt}} % -8
\newcommand{\cutcaptiondown}{\vspace*{-4pt}}

\newcommand{\cutequationup}{\vspace*{-4pt}}
\newcommand{\cutequationdown}{\vspace*{-0pt}}

\else
\newcommand{\cutsectionup}{\vspace*{-0pt}} 
\newcommand{\cutsectiondown}{\vspace*{-0pt}}
\newcommand{\cutsubsectionup}{\vspace*{-0pt}}
\newcommand{\cutsubsectiondown}{\vspace*{-0pt}}

\newcommand{\cutparagraphup}{\vspace*{-0pt}} % -10

\newcommand{\cutcaptionup}{\vspace*{-0pt}} % -9
\newcommand{\cutcaptiondown}{\vspace*{-0pt}}

\newcommand{\cutequationup}{\vspace*{-0pt}}
\newcommand{\cutequationdown}{\vspace*{-0pt}}

\fi

\def\paperID{8969} % *** Enter the Paper ID here
\def\confName{ICCV}
\def\confYear{2025}

\def\ourmodel{Orchid}

\title{Orchid: Image Latent Diffusion for Joint Appearance and Geometry Generation}
\author{Akshay Krishnan\textsuperscript{1,2,*}% \footnotemark[1]
\quad Xinchen Yan\textsuperscript{1} \quad Vincent Casser\textsuperscript{3} \quad Abhijit Kundu\textsuperscript{1}\\
\textsuperscript{1}Google DeepMind, \textsuperscript{2}Georgia Institute of Technology,
\textsuperscript{3}Waymo\\
{\tt\small \href{https://orchid3d.github.io}{https://orchid3d.github.io}}
}

\newcommand{\cmark}{\textcolor{ForestGreen}{\text{\ding{51}}}}
\newcommand{\xmark}{\textcolor{red}{\text{\ding{55}}}}

\begin{document}

% ------------- Uncomment for main paper, comment for supplm.---------------- %
\twocolumn[{%
\renewcommand\twocolumn[1][]{#1}%
\maketitle
\vspace{-2em}
\includegraphics[width=1.0\linewidth]{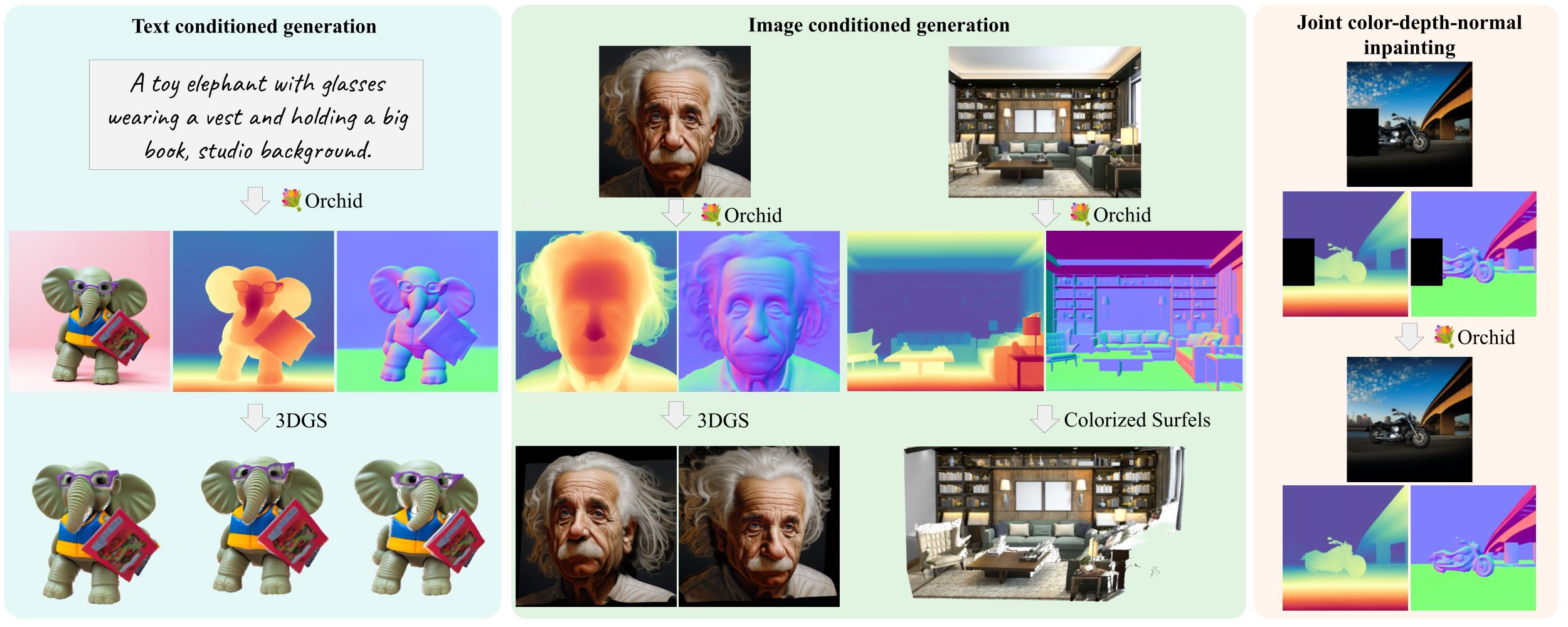}
% \cutcaptionup
% \cutcaptionup
\vspace{-2em}
\captionof{figure}{We propose \ourmodel{}: a unified, multi-modal latent diffusion model that jointly generates color, depth, and surface normals. The output color, depth, and normals are consistent with each other and can be seamlessly turned to 3D reconstructions using methods like 3DGS~\cite{kerbl3Dgaussians}. \ourmodel{} can generate 3D scenes from text (\textit{left}), or from a single color image (\textit{center}). Orchid captures joint appearance and geometry prior which can be used to solve different 3D inverse problems \eg inpaint incomplete 2.5D reconstructions (\textit{right}).}
\vspace{0.8em}  % was: 1.2em
%\cutcaptiondown
\label{fig:teaser}
}]
\renewcommand{\thefootnote}{\fnsymbol{footnote}}
\footnotetext[1]{Work done as an intern at Google DeepMind}
\renewcommand{\thefootnote}{\arabic{footnote}}

\begin{abstract}

%\cutcaptionup
\vspace*{-16pt}
We introduce Orchid, a unified latent diffusion model that learns a joint appearance-geometry prior to generate color, depth, and surface normal images in a single diffusion process. This unified approach is more efficient and coherent than current pipelines that use separate models for appearance and geometry. Orchid is versatile—it directly generates color, depth, and normal images from text, supports joint monocular depth and normal estimation with color-conditioned finetuning, and seamlessly inpaints large 3D regions by sampling from the joint distribution. It leverages a novel Variational Autoencoder (VAE) that jointly encodes RGB, relative depth, and surface normals into a shared latent space, combined with a latent diffusion model that denoises these latents. Our extensive experiments demonstrate that Orchid delivers competitive performance against SOTA task-specific methods for geometry prediction, even surpassing them in normal-prediction accuracy and depth-normal consistency. It also inpaints color-depth-normal images jointly, with more qualitative realism than existing multi-step methods.

\end{abstract}
\cutsectionup
\section{Introduction}
\label{sec:intro}

Imagine that you are planning to paint a beautiful picture of a café. You would not simply start painting—you would first sketch out the space, mapping the layout of seating areas, considering the slopes of the floor and walls, and capturing how light and shadow create the illusion of depth. This natural, integrated understanding of appearance and geometry underpins creative processes in almost every domain from architecture to video game development. Similarly, this joint prior of appearance and geometry is key for deep generative models to produce seamless 3D scenes, opening up transformative possibilities in VR~\cite{savva2019habitat,straub2019replica}, animation~\cite{saito2019pifu}, and robotics~\cite{gaidon2016virtual,abu2018augmented,chen2021geosim}. Although current 2D diffusion models excel at creating high-quality images from a rich appearance prior, we still face challenges in jointly modeling appearance and geometry, largely due to the limited scale of 3D datasets.

\begin{table}[t]
\begin{center}
\resizebox{\linewidth}{!}{
\begin{tabular}{l|c|c|c} 
\toprule
 & \multicolumn{2}{c|}{Multi-step generation} & Joint Diff \\ 
 & Diff + FF & Diff + Diff & (ours) \\ 
\midrule

\# Independent models & 3 & 3 & 1\\
Inference time (s / image) & 1.3* \includegraphics[width=1em]{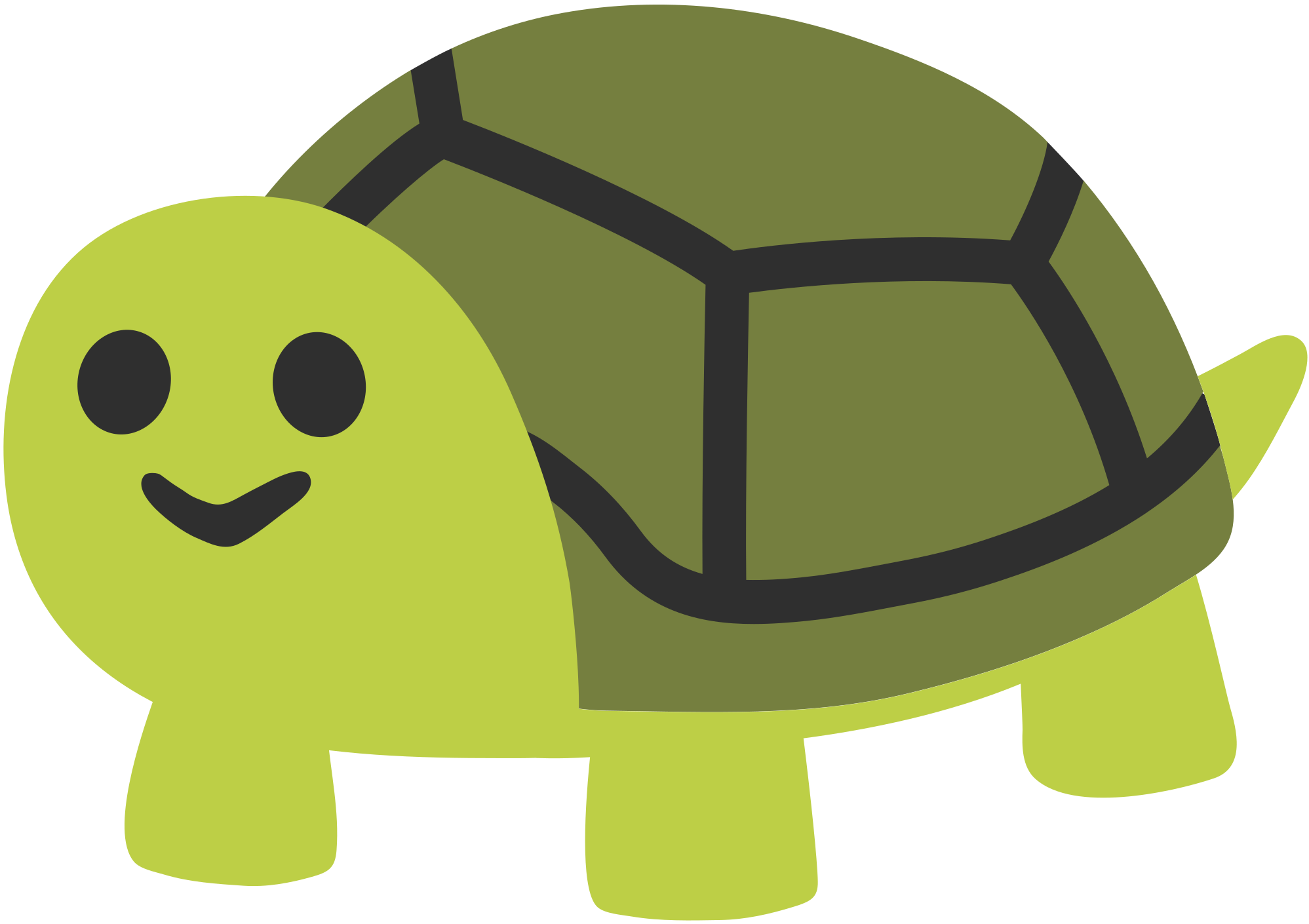}  & 4.2* \includegraphics[width=1em]{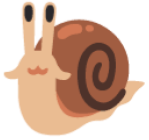} & 1.2 \includegraphics[width=1em]{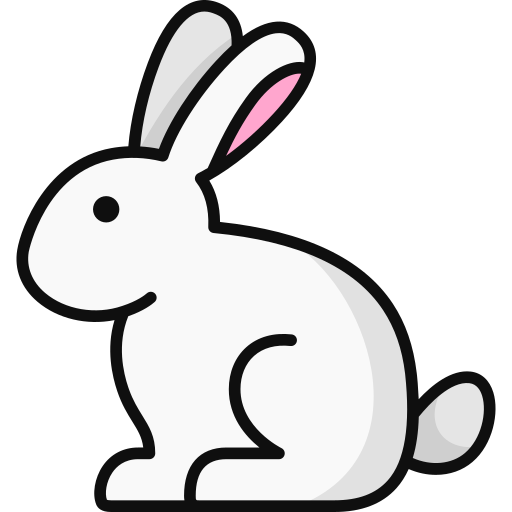} \\
Geometry generative prior & \includegraphics[width=1em]{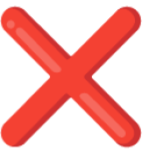} & \includegraphics[width=1em]{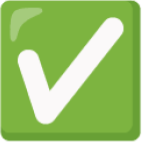} & \includegraphics[width=1em]{images/check.png} \\
Joint image-geometry prior & \includegraphics[width=1em]{images/cross.png} & \includegraphics[width=1em]{images/cross.png} & \includegraphics[width=1em]{images/check.png}\\
Consistent depth-normal &  \includegraphics[width=1em]{images/cross.png} & \includegraphics[width=1em]{images/cross.png} & \includegraphics[width=1em]{images/check.png}\\
\bottomrule
\end{tabular}
}
\end{center}
\cutcaptionup
\cutcaptionup
\caption{\textbf{Comparison of methodologies for 2.5D/3D generation:} \textit{Diff + FF} diffuses color image followed by feedforward (FF) models for depth + normals; \textit{Diff + Diff} diffuses color, then diffuses depth and normals conditioned on the color image; \textit{Joint Diff} (\ourmodel{}) jointly diffuses color, depth and normal. Methods with (*) cannot store all models on a GPU and need added I/O time not included here (more details in Appendix Section B).}
\cutcaptiondown
\cutcaptiondown
\label{tab:methods}
\end{table}

A viable option for the task of generating images along with its 3D geometry is to first generate a color image using a standalone color image generation model, followed by separate monocular depth and normal models. However, using separate task-specific models for depth and normal results in inconsistencies between their predictions. The overall generation system also becomes inefficient by using multiple large models, with high memory consumption and latency. In addition, this approach of first sampling from an appearance prior, and subsequently sampling from separate color-conditional depth and normal priors does not yield good results for inverse problems that entangle appearance and geometry - such as 3D inpainting, scene completion, or object manipulation. These problems require sampling from the joint space, given some partial observations (for example, a partial colored pointcloud). A method that first samples from a color-image prior using an inpainting model ignores the known scene geometry.

In this work, we propose ``\emph{\ourmodel{}}'': a unified latent diffusion model (LDM) for the joint generation of appearance and geometry. \ourmodel{} provides a joint generative latent prior that can be readily applied to various problems, avoiding the need for combinations of different task-specific priors. It comprises a new  Variational Autoencoder (VAE) to encode RGB, depth, and surface normals to a joint image latent space, and a text-conditioned LDM to denoise the joint latents. This design allows \ourmodel{} to address different 3D generation problems. Specifically, it can readily generate color, depth and normal images from text. It is the first model to jointly inpaint large 3D regions in the color-depth-normal image space. With color-conditioned finetuning, it can jointly predict  depth and normals from input color images, acting as a monocular geometry estimator. Both our VAE and our LDM leverage pretraining from large color image datasets. Additionally, we use multiple real world color-depth-normal datasets, along with distillation from discriminative teacher depth and normal models on a large text-image dataset.

As shown in Table \ref{tab:methods}, a joint LDM has several advantages over a combination of a color LDM and color-conditioned downstream models for depth or normals.
It generates color, depth, and normals in a single diffusion process, being faster and more memory efficient because only a single model is needed.
As shown by our experiments, its depth and surface normal predictions are more self-consistent than those produced by different state-of-the-art models. 
Finally, the joint prior can be directly applied to many 3D reconstruction problems: we show that it can be used to sample from the joint appearance-geometry distribution conditioned on partial input, such as during 3D scene inpainting. 
\ourmodel{} has several potential applications in robotics and virtual reality, such as image-conditioned depth completion for sparse LiDAR sensors or inpainting unseen regions during novel view synthesis from sparse views. It could also serve as a replacement for separate appearance-geometry models used in score-distillation frameworks such as~\cite{zhang2024nerfinpaintinggeometricdiffusion} by providing a unified appearance-geometry score function.

In summary, our contributions are as follows. 

\begin{itemize}
    \item We introduce \ourmodel{}, a unified latent diffusion model to generate color, depth and normals from text in a single diffusion process. We train a  new joint VAE to encode color, depth, and normals to a shared latent space. 
    \item We show that \ourmodel{} can act as a joint monocular depth and normal estimator with color-conditioned finetuning. Orchid beats state-of-the-art methods in terms of monocular normal prediction accuracy and consistency of depth and normal predictions. 
    \item \ourmodel{} learns a joint color-depth-normal prior that can be used for inverse problems. We show that is can consistently inpaint 3D scenes in the color-depth-normal space, producing qualitatively better results than a combination of color, depth and normal inpainting models.
\end{itemize}

\begin{figure*}[th]
    \centering
    \includegraphics[width=\linewidth]{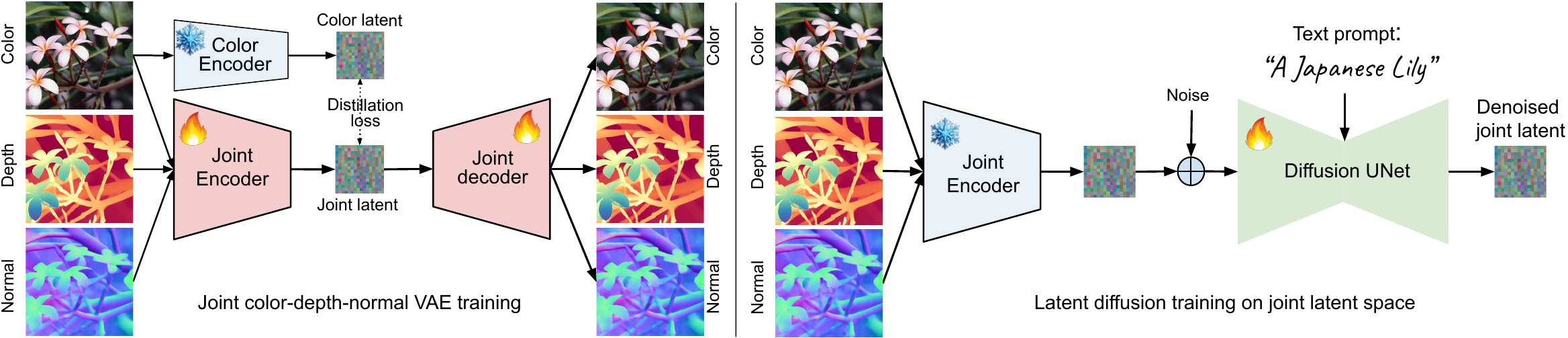}
    \cutcaptionup
    \cutcaptionup
    \caption
    {
    \textbf{\ourmodel{} training scheme:} As illustrated on the left figure, we first train a joint color-depth-normal VAE that leverages priors from color-only pretraining. 
    We introduce additional depth and normal reconstruction losses, along with a distillation loss to ensure that the joint latent space follows the structure of the original color-only VAE.
    As illustrated on the right figure, 
    we then train the latent diffusion model on paired image-depth-normal and text data, while keeping the joint VAE encoder frozen. 
    }
    \cutcaptiondown
\label{fig:vae_arch}
\end{figure*}

\section{Related work}
\label{sec:related_work}
\cutsectiondown

\paragraph{Monocular depth and normal prediction:} Modern deep depth prediction models rely on priors learned from large scale datasets~\cite{midas, li2018megadepth, Bhat_2021_CVPR, eigen2014depth, fu2018deep, marigold, fu2024geowizard, depth_anything_v1, depth_anything_v2, hu2024metric3d} to generalize beyond training distribution. 
Since metric depth is a function of the camera's intrinsics parameters, most approaches, including ours, predict affine-invariant depth. 
Models do that predict metric depth are either limited to a single camera \cite{eigen2014depth} or condition on the camera intrinsics \cite{Bhat_2021_CVPR, hu2024metric3d, guizilini2023towards}. 

\textit{Monocular surface normal prediction} has also seen similar trends, with feed-forward deep models being the standard approach \cite{wang2015designingnormal, eftekhar2021omnidata}. In particular, \cite{eftekhar2021omnidata} achieved state-of-the-art performance following this setup. The most recent state-of-the-art \cite{bae2024dsine} incorporates additional inductive biases specific to the normal prediction task into the model. 

\textit{Joint prediction of depth and surface normals} has been explored by a few works as a multi-task problem. They are modelled using two branches on a deep network with a shared backbone \cite{eigen2015predictingnormal, li2015depth, xu2018pad, zhang2019pattern}.

Most recent approaches for depth \cite{depth_anything_v1, depth_anything_v2} and normal prediction \cite{hu2024metric3d} adopt self-supervised ViT backbones \cite{oquab2023dinov2}, and train jointly on several large datasets to achieve zero-shot in-the-wild generalization. Although they make predictions from images, they are discriminative and do not learn a generative prior that can be leveraged for tasks like depth completion or refinement. 

\cutparagraphup
\cutparagraphup
\paragraph{Diffusion priors for depth and normals:}
Several recent works~\cite{marigold, fu2024geowizard, saxena2023monocular, saxena2024surprising, duan2023diffusiondepth, ji2023ddp, he2024lotus, du2023generative, zhang2025zerocomp} have shown that diffusion priors learned from large color image datasets can be adapted for depth and normal estimation by finetuning them on relatively small amounts of data.
These color-conditioned diffusion models learn a generative 2D prior which can be used for problems like depth completion. GeoWizard~\cite{fu2024geowizard} is an existing diffusion model for joint depth-normal prediction, but it cannot generate color. It also uses separate representations with ``switch conditioning'' for depth and normals, consuming more memory. 

When used to generate 3D scenes, these methods require multiple diffusion processes and models: first to generate color, then to generate depth and/or normals.
In contrast, our model learns a joint prior over color, depth and normal. 

\cutparagraphup
\cutparagraphup
\paragraph{Diffusion priors for inverse problems:} 2D text-to-color diffusion models trained on large datasets are emerging as promising generative priors for several applications. They are used as denoisers within an optimization framework, with other auxiliary objectives. They are leveraged for text-to-3D generation using a score-distillation sampling (SDS) loss to optimize a neural 3D representation \cite{poole2022dreamfusion,liu2023zero,lin2023magic3d,yan2024dreamdissector,wang2024prolificdreamer}. They can also be used for 3D object manipulation  \cite{wang2024diffusionmodelsgeometrycritics, wu2024neural}, depth completion \cite{viola2024marigolddc}, and 3D scene completion  \cite{weber2023nerfiller, shriram2024realmdreamer, prabhu2023inpaint, mirzaei2024reffusionreferenceadapteddiffusion, zhang2024nerfinpaintinggeometricdiffusion}. The biggest challenge for such methods is the multi-view inconsistency of color diffusion models, as inconsistent appearance predictions across views cause artifacts in 3D reconstructions. These methods only rely on appearance priors. Some works \cite{zhang2024nerfinpaintinggeometricdiffusion} combine color with geometric diffusion priors by balancing multiple diffusion models. Orchid introduces a stronger unified appearance-geometry prior in a single model.

\section{Method}
\begin{figure}[t]
    \centering
    \includegraphics[width=\linewidth]{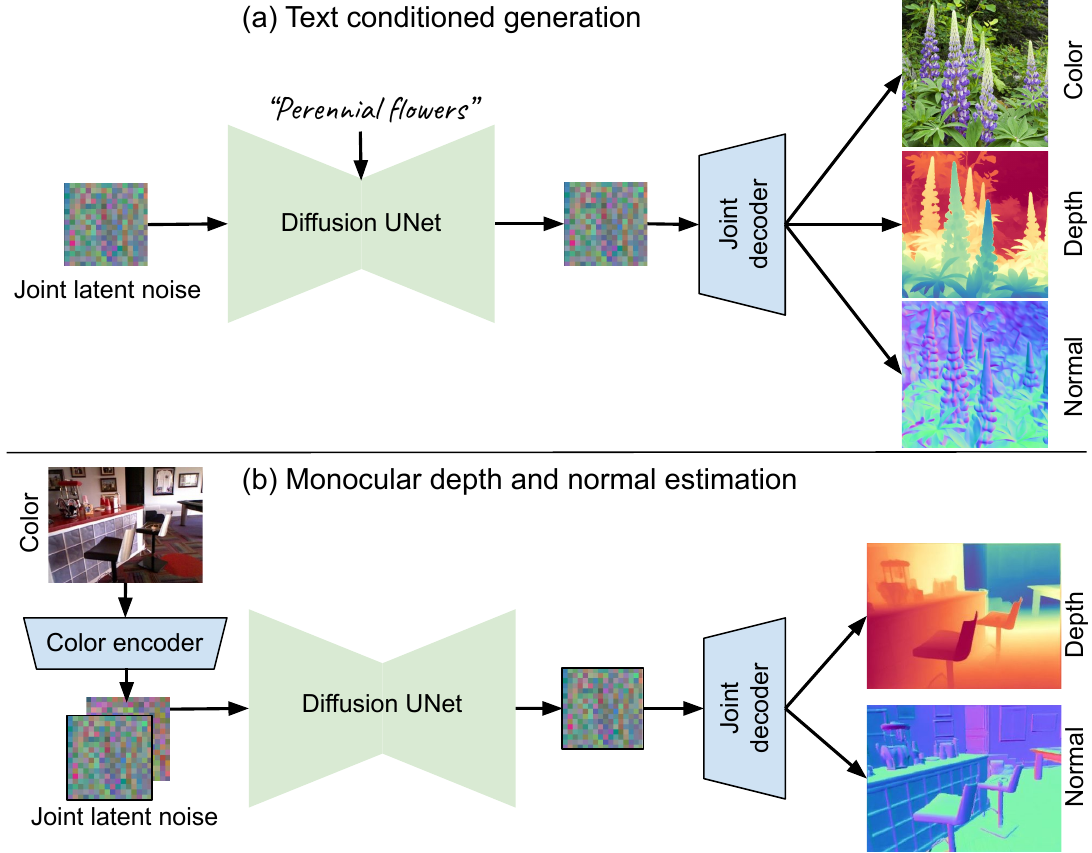}
    \cutcaptionup
    \cutcaptionup
    \caption
    {
    \textbf{\ourmodel{} inference:} (a) For text-conditioned generation, we denoise the joint latents conditioned on the text prompt; (b) For monocular depth / normal estimation, we denoise the joint latents with a noise-free input color-only latent as condition (no text).
    }
    \cutcaptiondown
\label{fig:arch_inference}
\end{figure}

We train \ourmodel{}, an LDM to learn a joint 2D generative prior of appearance (RGB) and geometry (depth and surface normals).
Specifically, it learns a score function for the joint distribution $p(\mathbf{x}, \mathbf{d}, \mathbf{n} | \mathcal{T})$ of color $\mathbf{x} \in \mathbb{R}^{H\times W\times 3}$, depth $\mathbf{d} \in \mathbb{R}^{H \times W}$, and surface normals $\mathbf{n} \in {[-1, 1]}^{H\times W\times 3}$ (with $||\mathbf{n}||_2 = 1$), conditioned on text prompts $\mathcal{T}$. This is a foundational generative prior that can be used for various tasks: text-conditioned appearance and geometry generation (sampling from $p(\mathbf{x}, \mathbf{d}, \mathbf{n} | \mathcal{T})$), color-conditioned depth and normal estimation (sampling from $p(\mathbf{d}, \mathbf{n} | \mathbf{x})$), and joint appearance-geometry inpainting (sampling from $p(\mathbf{x}, \mathbf{d}, \mathbf{n} | \mathcal{T}; \mathbf{x'}, \mathbf{d'}, \mathbf{n'})$ where  $\mathbf{x'}, \mathbf{d'}, \mathbf{n'}$ are partial observations). 
The ability to condition on partial observations in the joint space is unique to \ourmodel{}.

Central to Orchid is (1) a VAE (Section~\ref{sec:joint-vae}) that encodes color, depth and normals to a joint latent space, building upon priors learned by color-space VAE; and (2) a text-conditioned LDM (Section~\ref{sec:joint-diffusion}) with the ability to generate depth and normals in addition to color.
We also finetune \ourmodel{} for monocular depth and normal prediction, with an added color image condition (Section~\ref{sec:image-cond-depth-normal}).

\cutsubsectionup
\subsection{Joint latent space VAE}
\label{sec:joint-vae}
\cutsubsectiondown

The VAE encoder~\cite{kingma2013auto,rombach2022high} used by color LDMs \cite{rombach2021highresolution} projects RGB images to a compact latent space for iterative denoising.
We extend this VAE by adding four channels (one for depth and three for normals) to jointly encode color, depth and normals.
This lets \ourmodel{} decode all three modalities at once, unlike other task-specific LDMs~\cite{marigold, fu2024geowizard, he2024lotus} that use separate latents and LDMs for each modality. 
The joint latent space also minimizes redundancies in depth and normal representations. 
Moreover, training our VAE enables us to use pixel space representations that are best suited for geometry. For \eg, using unbounded inverse depth, which yields better predictions at longer ranges, unlike models that reuse a color-VAE with a [0-1] depth normalization using near-far planes. 

We preprocess depth before it is input to the VAE. In particular, from metric depth $\mathbf{d}$, we compute inverse depth  $\mathbf{d^*} = 1 / {\mathbf{d}}$, and its deviation around the median: $d_\sigma = \text{mean}(|\mathbf{d^*} - \text{median}(\mathbf{d^*})|)$. We  normalize inverse depth by the deviation $\mathbf{d'} = \mathbf{d^*} / d_\sigma$. We then shift it to begin at zero:  $\mathbf{d}_\text{model} = \mathbf{d'} - {d'}_\text{min}$. To simplify notation, we use $\mathbf{d}$ to denote the preprocessed inverse depth $\mathbf{d}_\text{model}$. 

Starting from a color-pretrained VAE, we train our joint VAE on paired color, depth, and normal datasets.
We apply losses for joint color, depth, and normals  ($L_\mathbf{x}$, $L_\mathbf{d}$, $L_\mathbf{n}$), along with a KL regularization loss $L_\text{KL}$ and a distillation loss $L_\text{distill}$ on the joint latent space.
The distillation loss is crucial to encourage our VAE's latents to be close in distribution to the color VAE, to retain diffusion priors from color pretraining (Section \ref{sub:ablations}).
Our color loss $L_\mathbf{x}$ contains a reconstruction loss $L_\text{rec} = || \mathbf{\hat{x}} - \mathbf{x} ||_2$, adversarial loss $L_\text{adv}$, perceptual loss $L_\text{LPIPS} = || \mathcal{F}_\text{VGG}(\mathbf{\hat{x}}) - \mathcal{F}_\text{VGG}(\mathbf{x}) ||_2$~\cite{zhang2018unreasonable}, and locally-discriminative learning loss $ L_\text{local\_disc}$~\cite{jie2022LDL}. 
These losses follow the standard practices from conventional color-space VAE  literature~\cite{rombach2022high}. 

We supervise our depth and normal prediction using depth reconstruction loss $L_\text{depth\_rec} = || \mathbf{\hat{d}} - \mathbf{d} ||_1$, the multi-resolution scale-invariant depth gradient loss $L_\text{depth\_grad}$  \cite{li2018megadepth}, and the normal reconstruction loss $L_\text{normal\_rec} = || \mathbf{\hat{n}} - \mathbf{n} ||_2$. 
The full training loop is summarized in Alg.~\ref{alg:vae_training}. We further use an exponential moving average for updating the VAE encoder and decoder parameters.

\cutequationup
\begin{align*}
L_\mathbf{x} &= w_1^\mathbf{x} L_\text{rec} + w_2^\mathbf{x} L_\text{adv} + w_3^\mathbf{x} L_\text{LPIPS} + w_4^\mathbf{x} L_\text{local\_disc}\\
L_\mathbf{d} &= w_1^\mathbf{d} L_\text{depth\_rec} + w_2^\mathbf{d} L_\text{depth\_grad}\\
L_\mathbf{n} &= w^\mathbf{n} L_\text{normal\_rec}
\end{align*}

\cutequationup
\begin{algorithm}
\caption{Joint VAE training}\label{alg:vae_training}
\begin{algorithmic}
\State Initialize model weights $\mathbf{\theta} = [\theta_\text{enc}, \theta_\text{dec}, \theta_\text{disc}]$
\For{$i = 1, \cdots ,\text{num\_steps}$}
\State Sample coupled color $\mathbf{x}$, depth $\mathbf{d}$, normal $\mathbf{n}$
\State $[\mathbf{z_{\mu}}, \mathbf{z_{\sigma}}] \gets \text{Enc}(\mathbf{x}, \mathbf{d}, \mathbf{n}; \theta_\text{enc})$
\State $\mathbf{z}_\text{sample}  \gets \mathcal{N}(\mathbf{z_{\mu}}, \mathbf{z_{\sigma}})$
\State $L_\text{KL} \gets w^\text{KL} * \text{KL}(\mathcal{N}(\mathbf{z_{\mu}}, \mathbf{z_{\sigma}}) \ || \  \mathcal{N}(\mathbf{0}, \mathbf{\Sigma}))$
\State $[\mathbf{z_{\mu}^{*}}, \mathbf{z_{\sigma}^{*}}] \gets \text{Enc}(\mathbf{x}, \mathbf{d}, \mathbf{n}; \theta_\text{enc}^{*})$
\State $L_\text{distill} \gets w^\text{distill} * || \mathbf{z_{\mu}^{*}} - \mathbf{z_{\mu}} ||_1 $
\State $[\mathbf{\hat{x}}, \mathbf{\hat{d}}, \mathbf{\hat{n}}] \gets \text{Dec}(\mathbf{z}_\text{sample}; \theta_\text{dec})$
\State $L_\text{disc} \gets \log(\text{Disc}(\mathbf{x}; \theta_\text{disc})) + \log(1 - \text{Disc}(\mathbf{\hat{x}}; \theta_\text{disc}))$
\State Update parameters according to gradients
\State $\theta_\text{enc} \stackrel{+}\gets -\nabla_{\theta_\text{enc}} (L_\mathbf{x} + L_\mathbf{d} + L_\mathbf{n} + L_\text{KL} + L_\text{distill})$
\State $\theta_\text{dec} \stackrel{+}\gets -\nabla_{\theta_\text{dec}} (L_\mathbf{x} + L_\mathbf{d} + L_\mathbf{n} - \gamma L_\text{disc})$
\State $\theta_\text{disc} \stackrel{+}\gets -\nabla_{\theta_\text{disc} } L_\text{disc} $
\EndFor
\end{algorithmic}
\end{algorithm}
\cutequationdown

\cutsubsectionup
\subsection{Joint diffusion prior for color-depth-normal}
\label{sec:joint-diffusion}
\cutsubsectiondown

We train a latent diffusion model (LDM) to jointly denoise color-depth-normal latents $\mathbf{z}$, conditioned on text embeddings (Figure \ref{fig:vae_arch}).
In the forward process, we inject noise in the joint latent space through $\mathbf{z}_t = \sqrt{\overline{\alpha_t}} \mathbf{z}_0 + \epsilon \sqrt{1-\overline{\alpha_t}}$ and $\epsilon \sim \mathcal{N}(0, \mathbf{I})$ by following $q(\mathbf{z}_t|\mathbf{z}_0) = \mathcal{N}(\mathbf{z}_t;  \sqrt{\overline{\alpha_t}} \mathbf{z}_0, (1-\overline{\alpha_t})\mathbf{I})$. Here, $\mathbf{z}_t$ represents the noisy input at diffusion time $t$ and $\overline{\alpha_t} = \prod_{s=1}^t (1 - \beta_s)$ is the diffusion coefficient~\cite{ho2020denoising}.
In the \textit{reverse} process, our latent diffusion model (parameterized by $\theta_\text{LDM}$) predicts the target $y$ given $\mathbf{z}_t$ and diffusion time $t$.
We optimize the standard objective function $L_\text{LDM} = \mathbb{E}_{\epsilon \in \mathcal{N}(0, I), t \in \mathcal{U}(T)}[\| y - f_{\theta_\text{LDM}}(\mathbf{z}_t, t) \|^2]$.
We use the $v$-prediction (velocity) parametrization for our target  $y$~\cite{salimans2022progressive}. 
This formulation allows us to directly use \ourmodel{} to generate color, depth, and normals from text prompts with classifier-free guidance \cite{ho2022classifierfreediffusionguidance} (Figure \ref{fig:arch_inference}a). 

\subsection{Color-conditioned depth and normal prediction}
\label{sec:image-cond-depth-normal}
\cutsubsectiondown
\begin{figure*}[t]
    \centering
    \includegraphics[width=\linewidth]{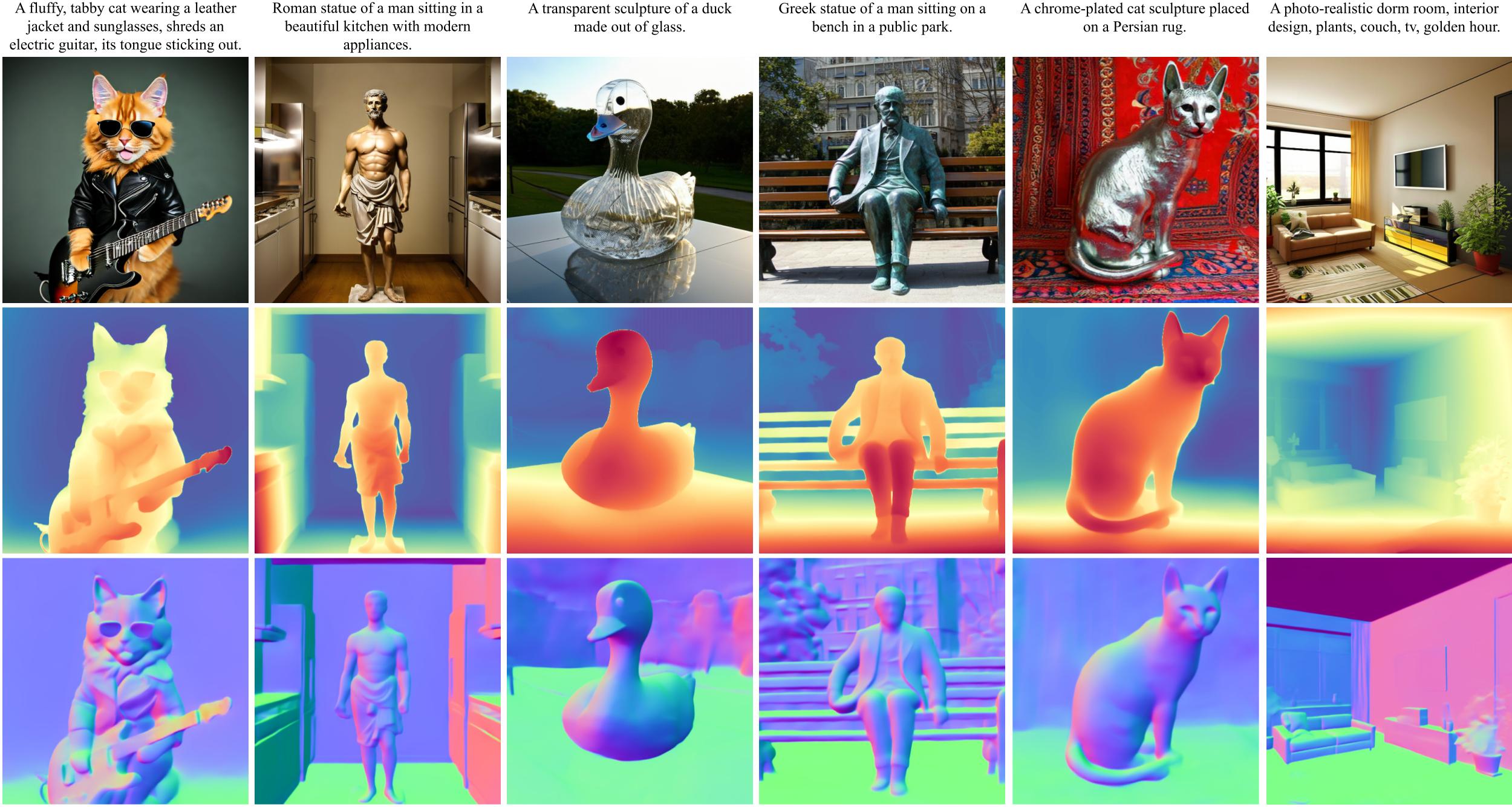}
    \cutcaptionup
    \cutcaptionup
    \caption
    {\textbf{Text-conditioned generation}: 
    Color, depth and normal images generated jointly by our model from a single text prompt. The input text prompt is provided on the top. 
    The generated color, depth, and normal are presented from top to bottom.
    Results show that \ourmodel{} can produce consistent appearance and geometry using its unified LDM for a wide range of text prompts such as close-up object shots or views from complex layouts of indoor and outdoor environment.
    }
    \cutcaptiondown
\label{fig:text_gen}
\end{figure*}

To further use \ourmodel{}'s learned color-depth-normal prior to jointly generate depth and surface normals from an input color image condition, we append an input color latent  $\mathbf{z}^\mathbf{x}$ from the pretrained frozen color-only VAE as a condition signal for diffusion through $f_{\theta_\text{LDM}}(\mathbf{z}_t, t; \mathbf{z}^\mathbf{x})$. While this is in line with previous work on using diffusion priors for dense prediction \cite{marigold, he2024lotus}, our joint latent space allows direct generation of \emph{both depth and surface normals} from color conditions. When used as an image-conditioned geometry estimator at inference, we drop text conditions (Figure \ref{fig:arch_inference}b). The sampling procedure is detailed in Alg.~\ref{alg:depth_sampling}. 

\cutequationup
\begin{algorithm}
\caption{Color-conditioned depth-normal sampling}\label{alg:depth_sampling}
\begin{algorithmic}
\Require{$\mathbf{z}^\mathbf{x}$ (color latent condition), $T$ (denoising steps)}
\State Initialize joint latent $\mathbf{z}_T \sim \mathcal{N}(\mathbf{0}, \mathbf{I})$
\For{$t=T, \dotsc, 1$}
    %\For{$u=1, \dotsc, U$}
    %\State $\epsilon \sim \mathcal{N}(\mathbf{0}, \mathbf{I})$ if $t > 1$, else $\epsilon = \mathbf{0}$
    %\State $\mathbf{z}_{t-1}^\text{known} = \sqrt{\bar{\alpha}_t} \mathbf{z}_0 + (1-\bar{\alpha}_t) \epsilon$
    \State $\eta \sim \mathcal{N}(\mathbf{0}, \mathbf{I})$ if $t > 1$, else $\eta = \mathbf{0}$
    \State $\mathbf{z}_{t-1} =
    \frac{1}{\sqrt{\alpha_t}}\left(\mathbf{z}_t  - \frac{\beta_t}{\sqrt{1-\bar\alpha_t}} f_{\theta_\text{LDM}}(\mathbf{z}_t, t; \mathbf{z}^\mathbf{x}) \right)
    + \sigma_t \eta$
\EndFor
\State \textbf{return} $\mathbf{z}_0$
\end{algorithmic}
\end{algorithm}
\cutequationdown
\cutequationdown

\begin{figure*}[t]
    \centering
    \includegraphics[width=\linewidth]{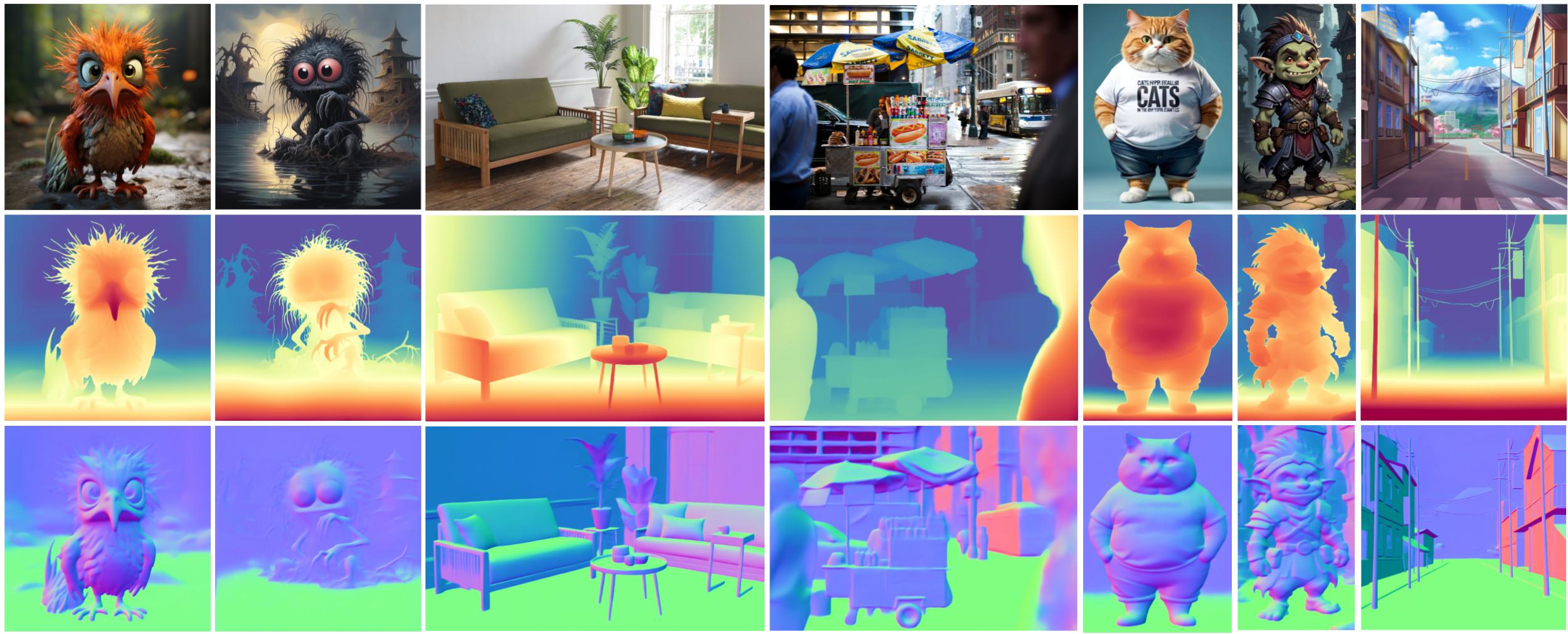}
    \cutcaptionup
    \cutcaptionup
    \caption
    {\textbf{Joint depth and surface normal prediction from single color image}: Given in-the-wild images as input (top row), \ourmodel{} works jointly predicts accurate and consistent depth (middle row) and normal (bottom row).
    }
    \cutcaptiondown
\label{fig:color_to_depth_normal}
\end{figure*}

% \cutsectionup
\section{Experiments}
\cutsectiondown

\ourmodel{} learns a joint color, depth, and surface normal prior that can be leveraged for several tasks. 
In our experiments, we evaluate its performance on three different tasks: text-conditioned color-depth-normal generation, color-conditioned depth and normal prediction, and unconditional joint color-depth-normal inpainting.
Our experiments highlight the versatility of \ourmodel{}, showing how a joint appearance and geometry prior can be used for either text or color conditioned generation, or as an unconditional model for solving inverse problems that combine appearance and geometry.

\cutsubsectionup
\subsection{Implementation details}\label{sub:implementation_details}
\cutsubsectiondown

We use a convolutional encoder and decoder with a latent dimension of 8 for our joint VAE, initialized from a VAE pretrained on datasets of color images alone. 
Our LDM is a transformer UNet similar to Stable Diffusion~\cite{rombach2022high}, pretrained on a large image-text dataset.
The diffusion model uses \textit{v}-prediction.
We train both models on 512x512 image resolution.
When training our text and color-conditioned diffusion models, we use a DDPM noise schedule~\cite{ho2020denoising} with 1000 steps. 
For diffusion sampling, we adopt a DDIM schedule with 100 steps for text-conditioned generation and 50 steps for color-conditioned generation. We use the zero terminal SNR schedule from \cite{lin2024commondiffusionnoiseschedules}. 

\textbf{Dataset:} We combine many real and synthetic datasets with ground truth depth and surface normals to train our VAE: Omnidata~\cite{eftekhar2021omnidata}, Virtual KITTI~\cite{gaidon2016virtual}, Hypersim~\cite{roberts:2021}, and DIODE~\cite{diode_dataset}. This amounts to about 2.5M images. Additionally, we also distill predictions from pretrained teacher depth~\cite{depth_anything_v2} and surface normal~\cite{hu2024metric3d} models on a text-image dataset with about 100M images. 

\cutsubsectionup
\subsection{Text-conditioned generation}
\cutsubsectiondown
\ourmodel{} is a first-of-its-kind model that can jointly generate color images along with their depth and normal from text prompts. 
We provide qualitative results of color, depth and normal generated by \ourmodel{} from a diverse set of text prompts in Figure~\ref{fig:text_gen}. 
Assuming approximate intrinsics, we can use the predicted color, depth and surface normal to optimize 3D Gaussians using Gaussian Splatting~\cite{kerbl3Dgaussians}, as shown in Figure~\ref{fig:teaser}.
More results (including a user study) are provided in the Appendix.
%
% More results are in supplementary material Section 4.2.

\begin{table*}[t]
\begin{center}
\resizebox{0.75\linewidth}{!}{
\begin{tabular}{l|c c c c c c c c}
\toprule
Model & \multicolumn{2}{c}{NYUv2} & \multicolumn{2}{c}{KITTI} & \multicolumn{2}{c}{ETH3D} & \multicolumn{2}{c}{ScanNet} \\ 
& AbsRel $\downarrow$ & $\delta_1$ $\uparrow$ & AbsRel $\downarrow$ & $\delta_1$ $\uparrow$ & AbsRel $\downarrow$ & $\delta_1$ $\uparrow$ & AbsRel $\downarrow$ & $\delta_1$ $\uparrow$ \\
\midrule
MiDaS~\cite{midas}                           & 11.7 & 87.5 & 23.6 & 63.0 & 18.4 & 75.2 & 12.1 & 84.6 \\
DepthAnything v2~\cite{depth_anything_v2}    & \cellcolor{tabfirst}\textbf{4.5}  & \cellcolor{tabfirst}\textbf{97.9} & \cellcolor{tabfirst}\textbf{7.4}  & \cellcolor{tabfirst}\textbf{94.6} & \cellcolor{tabthird}6.8  & 95.3 & \cellcolor{tabfirst}\textbf{6.0}  & \cellcolor{tabfirst}\textbf{96.3} \\
Marigold-depth~\cite{marigold}                     & 6.1  & 95.8 & \cellcolor{tabthird}9.8  & \cellcolor{tabthird}91.8 & \cellcolor{tabthird}6.8  & 95.6 & 6.9  & 94.6 \\
Lotus-G-depth \cite{he2024lotus}                                        & \cellcolor{tabsecond}5.4  & \cellcolor{tabthird}96.6 & 11.3 & 87.7 & \cellcolor{tabfirst}\textbf{6.2}  & \cellcolor{tabsecond}96.1 & \cellcolor{tabfirst}\textbf{6.0}  & \cellcolor{tabsecond}96.0\\
\hline
Geowizard~\cite{fu2024geowizard}             & \cellcolor{tabthird}5.6  & 96.3 & 14.4 & 82.0 & \cellcolor{tabsecond}6.6  & \cellcolor{tabthird}95.8 & \cellcolor{tabthird}6.4  & 95.0 \\
\ourmodel{} (Ours)                          & 5.7 & \cellcolor{tabsecond}96.9 & \cellcolor{tabsecond}7.7 & \cellcolor{tabsecond}94.4 & 7.3 & \cellcolor{tabfirst}\textbf{96.9} & \cellcolor{tabsecond}6.3 & \cellcolor{tabthird}95.8 \\
\bottomrule
\end{tabular}
}
\end{center}
\cutcaptionup
\cutcaptionup
\caption{\textbf{Zero-shot monocular depth estimation}: Comparison of our zero shot affine-invariant depth estimation accuracy to that of other methods. Being a joint depth-normal predictor, ours is comparable to the SOTA depth-only prediction model \cite{depth_anything_v2} while being better than the other joint depth-normal method \cite{fu2024geowizard}. The \colorbox{tabfirst}{first}, \colorbox{tabsecond}{second} and \colorbox{tabthird}{third} ranking methods are highlighted.}
\cutcaptiondown
\label{tab:image_to_depth}
\end{table*}

\begin{table*}[ht]
\begin{center}
\resizebox{0.75\linewidth}{!}{
\begin{tabular}{l| c c c c c c c c}
\toprule
Model & \multicolumn{2}{c}{NYUv2} & \multicolumn{2}{c}{ScanNet} & \multicolumn{2}{c}{iBims-1} & \multicolumn{2}{c}{Sintel} \\ 
& Mean $\downarrow$ & $11.25\degree$ $\uparrow$ & Mean $\downarrow$ & $11.25\degree$ $\uparrow$ & Mean $\downarrow$ & $11.25\degree$ $\uparrow$ &  Mean $\downarrow$ & $11.25\degree$ $\uparrow$\\ 
\midrule
Omnidata v2 \cite{eftekhar2021omnidata} & 17.2 & 55.5 & \cellcolor{tabthird}16.2 & 60.2 & 18.2 & 63.9 & 40.5 & 14.7\\
DSINE \cite{bae2024dsine} & \cellcolor{tabsecond}16.4 & \cellcolor{tabsecond}59.6 & \cellcolor{tabthird}16.2 & \cellcolor{tabthird}61.0 & \cellcolor{tabsecond}17.1 & \cellcolor{tabsecond}67.4 & \cellcolor{tabsecond}34.9 & \cellcolor{tabsecond}21.5\\
Marigold-normal \cite{marigold} & 20.9 & 50.5 & 21.3 & 45.6 & 18.5 & 64.7 & - & - \\
Lotus-G-normal \cite{he2024lotus} & \cellcolor{tabthird}16.9  & \cellcolor{tabthird}59.1 & \cellcolor{tabsecond}15.3 & \cellcolor{tabfirst}\textbf{64.0} & \cellcolor{tabthird}17.5 & \cellcolor{tabthird}66.1 & \cellcolor{tabthird}35.2 & \cellcolor{tabthird}19.9 \\
\hline
Geowizard \cite{fu2024geowizard} & 18.9 & 50.7 & 17.4 & 53.8 & 19.3 & 63.0 & 40.3 & 12.3\\
\ourmodel{} (Ours) & \cellcolor{tabfirst}\textbf{15.2} & \cellcolor{tabfirst}\textbf{60.6} & \cellcolor{tabfirst}\textbf{14.2} & \cellcolor{tabsecond}63.8 & \cellcolor{tabfirst}\textbf{16.3} & \cellcolor{tabfirst}\textbf{68.1} & \cellcolor{tabfirst}\textbf{31.7} & \cellcolor{tabfirst} \textbf{22.6}\\
\bottomrule
\end{tabular}
}
\end{center}
\cutcaptionup
\cutcaptionup
\caption{\textbf{Zero-shot monocular normals estimation}: \ourmodel{}'s color-conditioned surface normal predictions are significantly better than other SOTA methods, including specialized surface normal prediction models. \colorbox{tabfirst}{First}, \colorbox{tabsecond}{second} and \colorbox{tabthird}{third} ranking methods are highlighted.}
\cutcaptiondown
\cutcaptiondown
\label{tab:image_to_normal}
\end{table*}

\begin{figure}[t]
    \centering
    \includegraphics[width=\linewidth]{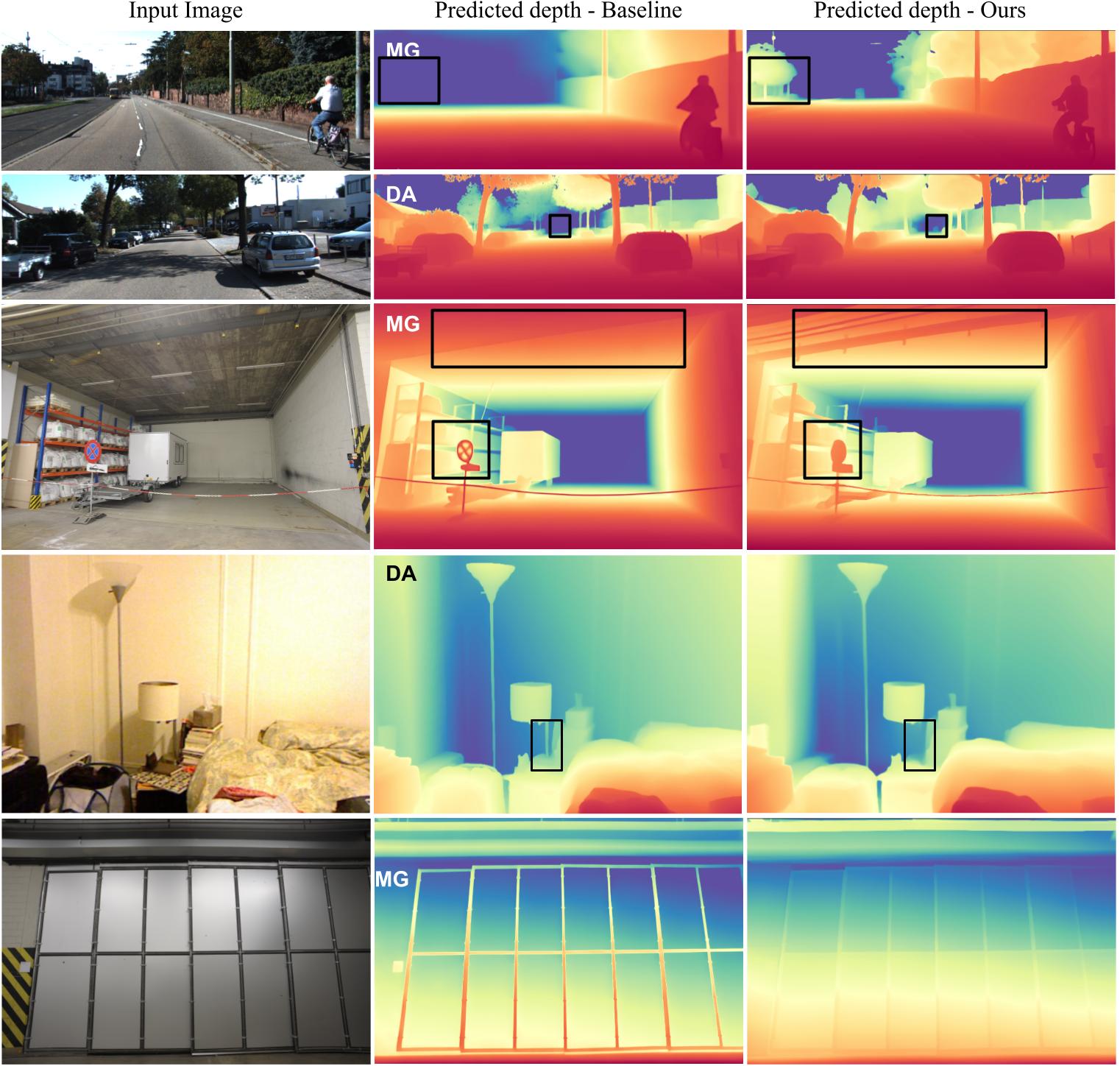}
    \cutcaptionup
    \cutcaptionup
    \caption
    {\textbf{Monocular depth prediction}: \ourmodel{} achieves clearly better performance than Marigold (MG)~\cite{marigold} on 
    far-range objects (row 1), thin structures (row 3), and depth-ambiguous scenes (row 5). It is comparable to Depth-Anything v2 (DA)~\cite{depth_anything_v2}, while being slightly better on very long range objects (row 2). Note that performance on far-away objects is not reflected in Table \ref{tab:image_to_depth} due to ground truth range limits.    
    }
    \cutcaptiondown
    \cutcaptiondown
\label{fig:mono_depth_vs_marigold}
\end{figure}
\begin{figure}[t]
    \centering
    \includegraphics[width=\linewidth]{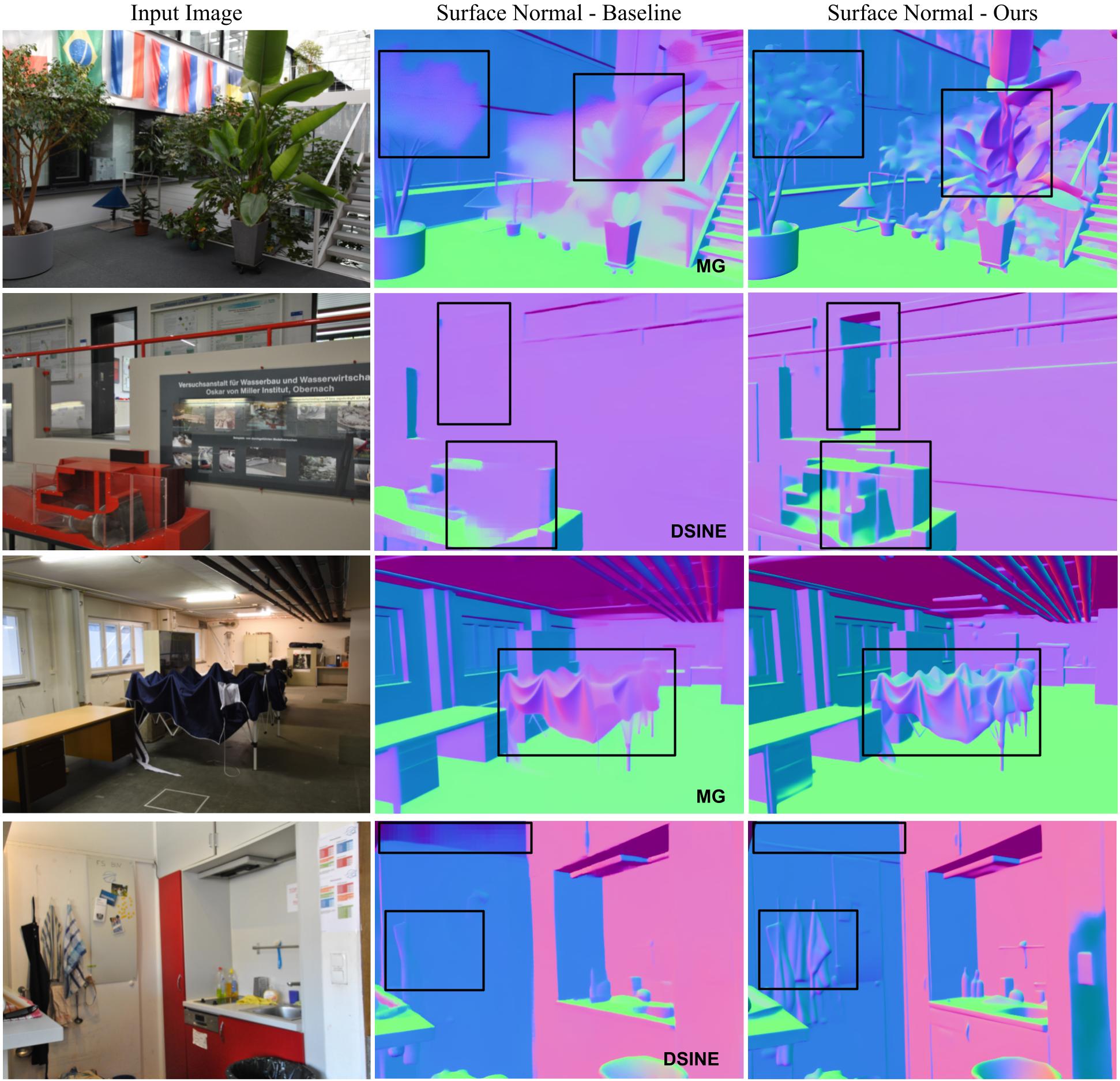}
    \cutcaptionup
    \cutcaptionup
    \caption
    {\textbf{Monocular surface normal prediction}: A comparison of our jointly generated normal to that of Marigold~\cite{marigold} and DSINE~\cite{bae2024dsine}. 
    \ourmodel{} produces more robust results on curved and deformable surfaces.
    }
    \cutcaptiondown
    \cutcaptiondown
\label{fig:monocular_normal_ours_vs_marigold}
\end{figure}
\begin{table}[ht]
\begin{center}
\vspace{1mm}
\resizebox{\linewidth}{!}{
\begin{tabular}{c | c c c c}
\toprule
Dataset & DA + DSINE & Marigold~\cite{marigold}& GeoWizard~\cite{fu2024geowizard} & \ourmodel{} \\
\midrule
NYUv2 & 0.059 & 0.102 & 0.122 & \textbf{0.040}\\
KITTI & 0.144 & 0.146 & 0.324 & \textbf{0.082} \\
\bottomrule
\end{tabular}
}
\end{center}
\cutcaptionup
\cutcaptionup
\caption{\textbf{Depth-normal inconsistency}: We show the mean error $e_\text{depth\_normal}$ ($\downarrow$) in the table. \ourmodel{}'s color-conditioned joint depth and normal estimates are significantly more consistent than those produced by GeoWizard, separate Marigold models, or a combination of Depth-Anything V2 and DSINE.
}
% \cutcaptiondown
\label{tab:depth_normal_consistency}
\end{table}
 
\begin{figure}[ht]
    \centering
    \includegraphics[width=0.9\linewidth]{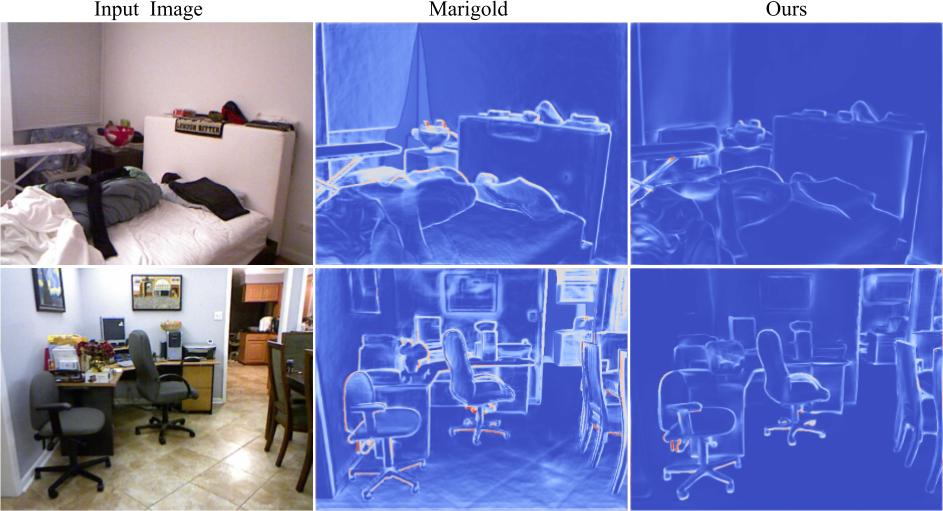}
    \cutcaptionup
    \caption
    {\textbf{Depth-normal consistency:} This heatmap of predicted depth-normal \emph{inconsistency} shows that our joint predictions are more consistent with each other, compared to Marigold~\cite{marigold}. 
    }
    \cutcaptiondown
\label{fig:depth_normal_consistency}
\cutcaptiondown
\end{figure}

\begin{figure*}[h]
    \centering
    \includegraphics[width=\linewidth]{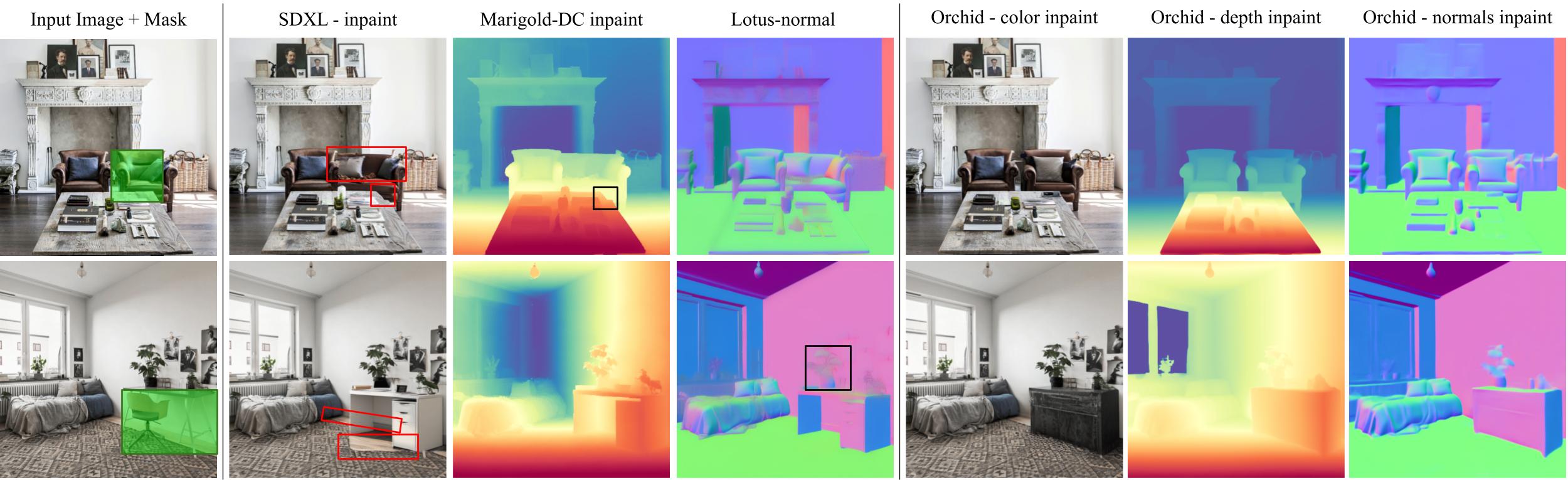}
    \cutcaptionup
    \cutcaptionup
    \caption
    {\textbf{Joint color-depth-normal inpainting}: Given a masked region in color-depth-normal data (left column shows masked color), our model inpaints them jointly, leveraging appearance and geometry cues. The baseline uses 3 diffusion models to first inpaint color, then inpaint depth and normals conditioned on it. Unlike ours, it produces geometric inconsistencies \eg discontinuous edges.
    }
    \cutcaptiondown
    \cutcaptiondown
\label{fig:inpainting}
\end{figure*}

\begin{table}[h]
\begin{center}
\vspace{1mm}
\resizebox{\linewidth}{!}{
\begin{tabular}{ l | c c | c c }
\toprule
Model &  \multicolumn{2}{c|}{Text-conditioned} & \multicolumn{2}{c}{Image-conditioned}\\
 & CLIP ($\uparrow$) & LPIPS ($\uparrow$) & $\delta_1$ ($\uparrow$) & $11.25\degree$ ($\uparrow$)\\
\midrule
\textbf{\ourmodel{}} & \textbf{0.316} & 0.764 & \textbf{96.9} & \textbf{60.6} \\
w/o joint latents & 0.312  & \textbf{0.769} &  96.0 & 54.8 \\
w/o distillation loss & 0.309  & 0.752 &  96.1 & 54.6 \\
w/o pseudo labels & - & - & 96.6 &  57.9 \\
\bottomrule
\end{tabular}
}
\end{center}
\cutcaptionup
\cutcaptionup
\caption{
\textbf{Ablations}: We find that our choices of joint latents, distillation loss, and depth-normal pseudo-labels are crucial. 
}
\cutcaptiondown
\cutcaptiondown
\label{tab:ablations}
\end{table}

% \cutsubsectionup
\subsection{Monocular depth and normal prediction}\label{sub:mono_depth_normal_details}
\cutsubsectiondown

We finetune the text-conditioned \ourmodel{} for color-conditioned prediction as explained in Section \ref{sec:image-cond-depth-normal}).
We evaluate the accuracy of depth and normals \emph{jointly estimated} by \ourmodel{} from color images. 
We follow the protocol in recent approaches and evaluate its zero-shot performance on depth and normal benchmarks we do not train on.
In addition, we present qualitative results on web images in Figure~\ref{fig:color_to_depth_normal}.
As illustrated in the figure, our model can produce high-fidelity and consistent depth and normals from in-the-wild color images from different domains: single objects, and complex indoor and outdoor outdoor scenes.

\cutparagraphup
\paragraph{Zero-shot monocular depth estimation:}
We evaluate the accuracy of the affine-invariant depth estimated by \ourmodel{} on four held-out datasets: NYUDepthv2~\cite{Silberman:ECCV12}, KITTI~\cite{Geiger2012CVPR}, ETH3D~\cite{schops2017multi} and ScanNet~\cite{dai2017scannet}, against other zero-shot depth estimation baselines.
This includes feed-forward~\cite{midas, hu2024metric3d, depth_anything_v2} as well as diffusion-based models~\cite{marigold, fu2024geowizard}. 
All baselines (except \cite{fu2024geowizard}) are trained solely for depth prediction while \ourmodel{} jointly predicts depth and normals.
From Table~\ref{tab:image_to_depth}, \ourmodel{}'s depth estimates are comparable to the state-of-the-art method~\cite{depth_anything_v2}, and better than other diffusion-based baselines.
Notably, \ourmodel{} achieves better monocular depth estimation performance than Geowizard~\cite{fu2024geowizard}, the only baseline that jointly predicts depth and surface normals. 
Figure~\ref{fig:mono_depth_vs_marigold} illustrates that our method produces more accurate depth than Marigold~\cite{marigold} on far-range objects, small objects, and depth-ambiguous scenes.
These benchmarks do not reflect performance on far-away objects on outdoor datasets like KITTI due to limited ground truth range, where \ourmodel{}'s predictions are qualitatively better. 

\cutparagraphup
\paragraph{Zero-shot monocular surface normals estimation:}
We evaluate \ourmodel{}'s normal estimation accuracy on four held-out datasets: NYUv2~\cite{Silberman:ECCV12}, ScanNet~\cite{dai2017scannet}, iBims-1~\cite{koch2018evaluation} and Sintel~\cite{butler2012naturalistic}.
From Table~\ref{tab:image_to_normal}, \ourmodel{} is significantly better than baselines at estimating surface normals.
This highlights the significance of training a VAE that explicitly encodes color, depth, and normals on a large dataset, as opposed to other diffusion baselines~\cite{marigold, fu2024geowizard, he2024lotus} that re-use a frozen color-space VAE for normals.
We qualitatively compare to Marigold-normal~\cite{marigold} and DSINE~\cite{bae2024dsine} in Figure~\ref{fig:monocular_normal_ours_vs_marigold}, where our method has better normal predictions, especially on curved and deformable surfaces.

\cutparagraphup
\paragraph{Depth-normal consistency:}
We evaluate the consistency of the depth and surface normal produced by \ourmodel{} and compare it to other baselines.
% that either use separate model weights or switch embedding to predict depth and normal~\cite{marigold, fu2024geowizard}.
%
We align affine-invariant depth $\mathbf{\hat{d}}$ with ground truth to obtain metric depth $\mathbf{d}$, and compute a 3D pointcloud $\mathbf{p}$ using camera intrinsics.
We estimate normals $\mathbf{\hat{n}} = \nabla_x \mathbf{p} \times \nabla_y \mathbf{p}$ from the pointcloud, and compute its inconsistency $e_\text{depth\_normal} = (1 - \mathbf{\hat{n}} \cdot \mathbf{n}) / 2$ with the estimated normals $\mathbf{n}$.
As shown in Table~\ref{tab:depth_normal_consistency}, we find that \ourmodel{} is able to leverage the joint latent space to predict depth and normal that are significantly more consistent. 
We visualize the depth-normal inconsistency error map in Figure~\ref{fig:depth_normal_consistency}.
These results clearly demonstrate the benefit of our joint diffusion model in downstream applications.

% \cutsubsectionup
\subsection{Joint color-depth-normal inpainting}\label{sub:inpainting}
\cutsubsectiondown

A unique ability of \ourmodel{}'s joint color-depth-normal diffusion prior is sampling from the joint distribution conditioned on partial observations, such as jointly inpainting color-depth-normal images.  Given color, depth and normals for a frame, we mask all of them for the region to be inpainted. Following the approach from RePaint \cite{lugmayr2022repaint}, we use \ourmodel{} as an unconditional inpainting prior to generate inpainted regions, as shown in Figure \ref{fig:inpainting}. We find that our model is able to generate consistent and realistic completions, without any additional training. A baseline inspired from existing 3D inpainting work~\cite{shriram2024realmdreamer, weber2023nerfiller, mirzaei2024reffusionreferenceadapteddiffusion} is to first run a Stable-Diffusion~\cite{rombach2022high} inpainting model, followed by depth inpatining \cite{viola2024marigolddc} and normal prediction \cite{he2024lotus}. Figure \ref{fig:inpainting} shows that this approach is clearly worse, as the color inpainting model produces geometric inconsistencies that are inherited by the geometry prediction models. Please refer to the Appendix for more results. 

% \cutsubsectionup
\subsection{Ablations}\label{sub:ablations}
% \cutsubsectiondown

We find some of our model and training design choices to be crucial to \ourmodel{}'s performance, as shown in Table \ref{tab:ablations}:
%

% Please find additional ablations in Appendix \ref{supp:ablations}.
\noindent \textbf{Pseudo depth-normal labels:} We use a large text-image dataset with depth and normals derived from pretrained teacher models. We find that this significantly boosts the accuracy of generated depth and normals, as measured on color-conditioned prediction. This is consistent with previous depth  models that use teacher predictions \cite{gui2024depthfm, depth_anything_v1}.

\noindent \textbf{Joint latent space:} We train an alternative unified color-depth-normal diffusion model that uses separate disentangled latents through a shared diffusion UNet, with an increased input/output dimension to accomodate all three latents. We find this to be significantly worse in terms of text-adherance as seen by a worse CLIP similarity. This is likely due to the joint distribution of disentangled latents drifting significantly from the LDM's color-pretrained latent space.

\noindent \textbf{Distillation loss:} It is used during VAE training to encourage our VAE's latents to be close in distribution to the original LDM's latents. We find that dropping it also causes a degradation in the generated images and geometry. 

% To summarize, our ablations show that the best way to learn a joint color-depth-normal prior is to learn a joint latent space that is of the same dimensionality and of similar structure to a large-scale color-image latent prior. Predictions from teacher depth/normal models significantly help learning this prior. 

% it makes the diffusion training forget some priors learned from image pretraining, generating qualitatively worse images. We observe a reduced CLIP image-text similarity by $0.7$ and increased FID by $1.7$. 

%\cutsectionup
\section{Conclusion}
% \cutsectiondown

Our work introduces \ourmodel{}, a novel joint appearance and geometry diffusion prior that encodes color, depth, and surface normals in a unified latent space. The joint appearance and geometry prior makes \ourmodel{} well-suited to various 3D reconstruction applications.
Notably, \ourmodel{} enables joint generation of images, depth, and surface normal from a text prompt in a single diffusion process.
%In particular, we show that it can be used for joint generation of images, depth and surface normals from an input text prompt.
%
Sampling from Orchid with a color image condition produces accurate and consistent depth and normals, which rival SOTA monocular depth and normal prediction models trained specifically for those tasks.
It also excels in joint inpainting of color, depth, and surface normals when used as an unconditional diffusion prior, a capability that is unique to \ourmodel{}. 
We anticipate \ourmodel{} will pave the way for advancements in tasks like new-view synthesis from single and sparse observations, densification from sparse depth, and solving inverse problems that entangle appearance and geometry.

\vspace{6mm}
\small{
\noindent\textbf{Acknowledgements:} We are very grateful to James Hays for his encouragement and feedback. We also thank Kyle Genova, Songyou Peng, Thomas Funkhouser, and Leonidas Guibas for their valuable insights during our discussions.
}
\clearpage

% ------------- Comment for main paper, uncomment for supplm.---------------- %
% Uncomment to arxiv
\appendix
\twocolumn[
\centering
\Large
\textbf{\thetitle}\\
\vspace{0.5em}Appendix \\
\vspace{1.0em}
] %< twocolumn

% Uncomment for supplemental
% \clearpage
% \setcounter{page}{1}
% \maketitlesupplementary

In this appendix, we provide additional details of our datasets, model architecture, ablations, and training methodology. We also provide a runtime analysis, user-study, and additional qualitative results from \ourmodel{} for text conditioned color-depth-normal generation as well as image conditioned depth-normal prediction and joint inpainting tasks, including comparisons to more baselines. We conclude with a discussion of our limitations, and scope for future work. Novel-view synthesis videos of 3D reconstructions using predictions from \ourmodel{} are provided on the web page \textit{\href{https://orchid3d.github.io}{https://orchid3d.github.io}}, along with a discussion in Section \ref{supp:3d_recon}. 

\section{\ourmodel{} details}

\subsection{Architecture}

For our VAE, we use a convolutional encoder and decoder with a latent dimension of 8, with $8\times$ spatial downsampling. The VAE has 7 input channels: 3 for RGB, 1 for depth, and 3 for surface normals. The discriminator (used only during VAE training) is a small ConvNet + MLP. 

Once the VAE is trained, we keep it frozen when training the latent diffusion model. The latent diffusion model itself is a UNet transformer similar to \textit{Stable Diffusion}~\cite{rombach2021highresolution} which is conditioned on both time and text embeddings. It has approximately 2B parameters. 

\subsection{Training}

We use a combination of RGB, depth, and normal losses when training the VAE, with weights as explained in Section 3.1 of our paper. Here, we provide the values of the weights we used for our model. For $L_\mathbf{x}$, we use $w^\mathbf{x}_1 = 1, w^\mathbf{x}_2 = 0.1, w^\mathbf{x}_3 = 0.1, w^\mathbf{x}_4 = 1$. For $L_\mathbf{d}$, we use $w^\mathbf{d}_1 = 1, w^\mathbf{d}_2  = 0.5$. We use $w^\mathbf{n} = 1$ for $L_\mathbf{n}$. We also use $w^{distill} = 10^-6$ for $L_{distill}$ and $w^{KL} = 10^-3$ for $L_{KL}$. Our choice of loss components and their weights for $L_\mathbf{x}$ and $L_\mathbf{KL}$ are based on standard training recipes for VAEs used in latent diffusion models. For losses we introduce, i.e, $L_\mathbf{d}$, $L_\mathbf{n}$, and $L_{distill}$, we obtained similar results with weights of similar orders of magnitude, but dropping them completely worsens quality (as shown in our ablations). 

On 16 NVIDIA A100 GPUs, we take approximately 5 days to train the VAE, 2 days to finetune our LDM starting from a color LDM, and 8-12 hours to finetune our image-conditioned model. 

\subsection{Dataset construction}

\begin{table}[h!]
\begin{center}
\resizebox{\linewidth}{!}{
\begin{tabular}{l | c c c c c} 
\toprule
Dataset & Size & Text & Depth & Normals \\ 
\midrule
Hypersim & 60k & \xmark & \cmark & \cmark \\
Virtual KITTI & 21k & \xmark & \cmark & \xmark \\
Replica + GSO (Omnidata) & 100k & \xmark & \cmark & \cmark \\
Taskonomy (Omnidata) & 2M &  \xmark & \cmark & \cmark \\
DIODE & 25k & \xmark & \cmark & \cmark \\
Pseudo-labeled (ours) & 110M & \cmark & \cmark & \cmark \\
\bottomrule
\end{tabular}
}
\end{center}
\cutcaptionup
\cutcaptionup
\caption{\textbf{Dataset details:} We use all  the above datasets for training the VAE, but only the pseudo-labeled text-image dataset, Hypersim, and Replica + GSO for finetuning our LDM.}
\cutcaptiondown
\label{tab:dataset}
\end{table}

We provide details of the dataset we use for VAE and LDM training in Table \ref{tab:dataset}. When training the VAE, we sample more heavily from the high-quality real world datasets, rather than our dataset with teacher model predictions. Whereas for the text-conditional LDM training, we sample more heavily from the distillation dataset which contains text-captions. For image-conditioned LDM finetuning, we ignore the text captions, and sample from both real-world and distillation data. While predictions from teacher models are not perfect, models distilled from multiple teachers have performed better in previous work\cite{wang2024samclipmergingvisionfoundation}. We remove a few rare examples where depth and normal teacher models disgree (high depth-normal inconsistency) for significant parts of the image.

\section{Runtime analysis}\label{supp:ablations}

\begin{table}[ht]
\begin{center}

\resizebox{\linewidth}{!}{
\begin{tabular}{c | c c c}
\toprule
Model & Diff + Diff & Diff + FF & Orchid \\  
\midrule
Inference time (s / img) & 4.2* & 1.3* & \textbf{1.2} \\
\bottomrule
\end{tabular}
}
\end{center}
\cutcaptionup
\cutcaptionup
\caption{\small \textbf{Runtime analysis:} Orchid is the fastest way to generate color, depth, and normals. A fair runtime comparison is hard since these methods vary in memory usage. Baselines using multiple models (*) cannot store all models on a GPU and need added weight I/O time that is not included here.}
\cutcaptiondown
\cutcaptiondown
\label{tab:supp_runtime}
\end{table}

We provide an analysis of the runtime taken for the different approaches discussed in Table 1 in our main paper in Table \ref{tab:supp_runtime}. We report inference times for on a single H100 for all three methods. Our joint generation of color, depth, and normals is significantly faster than generating them with 3 different diffusion models. It is also faster than using discriminative models for depth and normals after an image diffusion process - 1.2 vs 1.3 s per image. Although this difference may seem less significant, please note that we do not include the time taken to move model parameters to/from the GPU, which is required when using multiple models. This I/O time is significantly greater than the inference time for discriminative models.

\section{Ablation details}\label{supp:ablations}

This section provides details for some of the ablations provided in our paper. 

\noindent \textbf{Unified appearance-geometry diffusion baseline with disentangled latents:} \ourmodel{} uses a unified joint latent space for color-depth-normal generation. An alternative design to enable a unified color-depth-normal diffusion model would be to explicitly encode all three modalities using separate latents (all produced by the same VAE), and finetune the LDM to denoise a higher dimensional concatenation of all three latents. We find that while this is a feasible approach, the quality of generated images is significantly worse than that of using a joint latent. Our hypothesis is that this is likely due to a significant mismatch of the latent space from the color image-only pretraining stage, as opposed to a joint latent space that is similar in structure (due to the distillation loss) and dimensionality to the pretrained LDM's latent space. Quantitatively, Table 5 in our paper shows that this disentangled latents model has a lower CLIP-similarity score when evaluated on COCO captions. It does however have a slightly higher LPIPS, likely because it uses the same latent dimension to store color information alone. Our joint latent however is significantly better on image-conditioned prediction tasks, indicating that the model is able to learn an effective joint latent representation of all three modalities.

\begin{figure}[h]
    \centering
    \includegraphics[width=\linewidth]{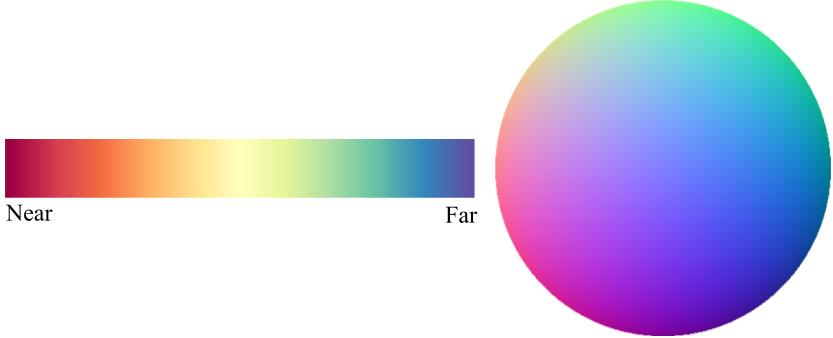}
    \cutcaptionup
    \cutcaptionup
    \caption
    {Colormap for depth (left) and surface normal on a unit hemisphere (right) used for all qualitative results in this paper. 
    }
    \cutcaptiondown
\label{fig:legend}
\end{figure}

\begin{figure*}[t]
    \centering
    \includegraphics[width=0.85\linewidth]{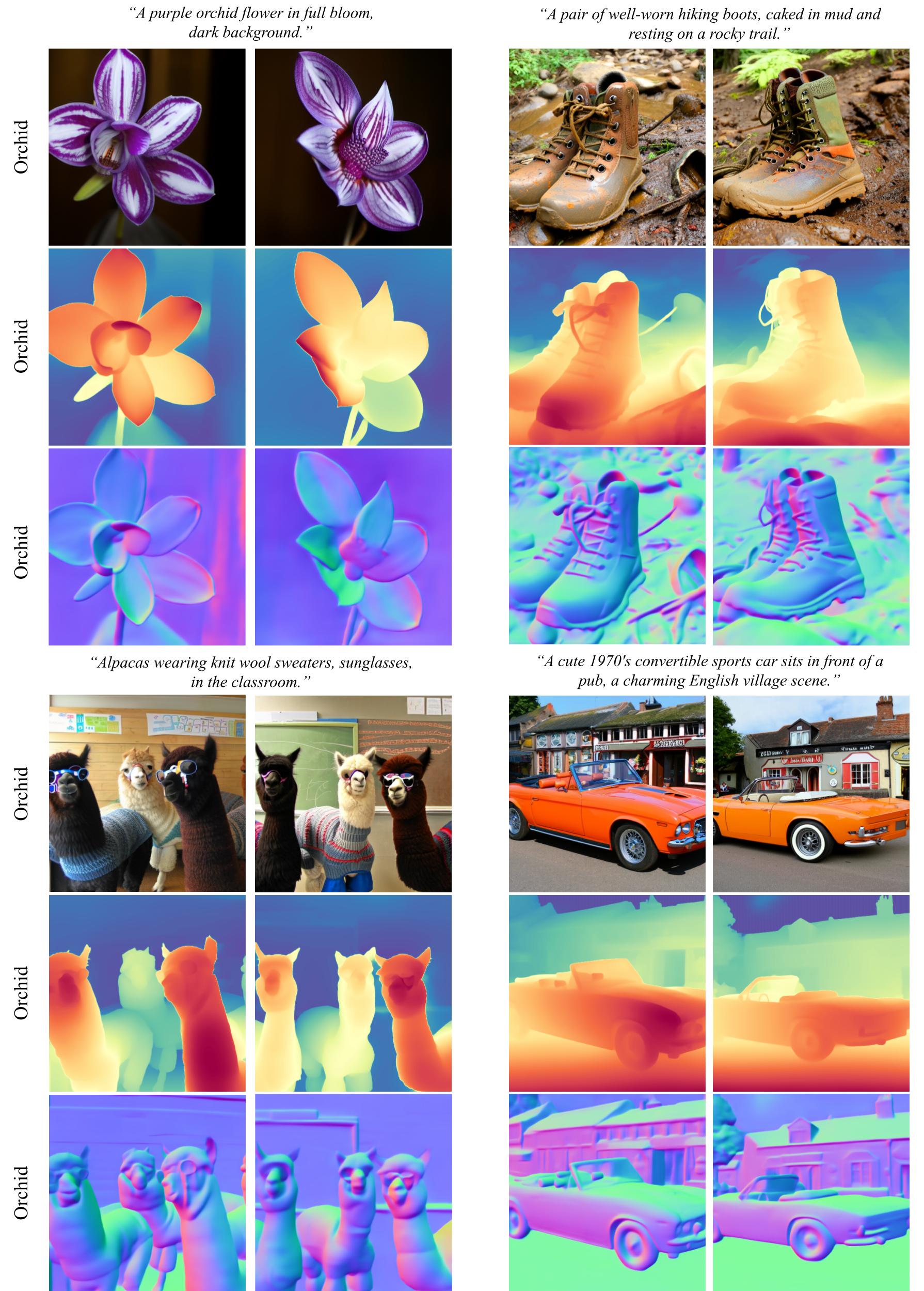}
    % \cutcaptionup
    \cutcaptionup
    \caption
    {\textbf{Text conditioned generation}: We show color, depth and normals generated by \ourmodel{} for different text prompts. We show two results for each prompt. 
    }
    \cutcaptiondown
\label{fig:supp_text_cond_ours}
\end{figure*}
\begin{figure*}[t]
    \centering
    \includegraphics[width=0.85\linewidth]{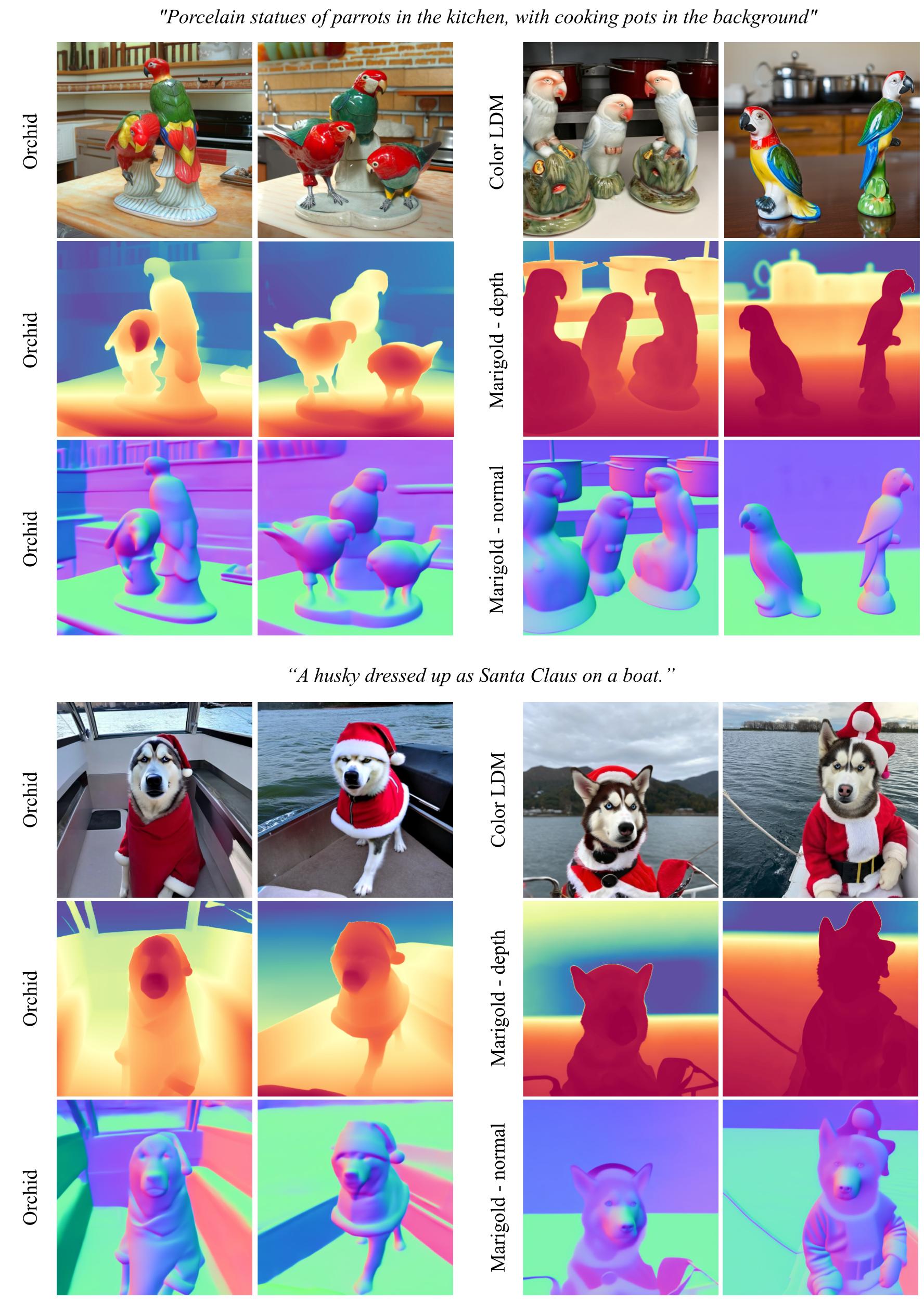}
    % \cutcaptionup
    \cutcaptionup
    \caption
    {\textbf{Text conditioned color-depth-normal generation:} We show two predictions from \ourmodel{} for each text prompt. We qualitatively compare these to the alternative: generate color, depth and normals from a separate diffusion model for each. For this baseline, we use a color-only LDM for color, and separate Marigold \cite{marigold} models for depth and normals. When comparing results, please refer to our note on depth map visualization (Section~\ref{sec:note_normalization}).
    }
    \cutcaptiondown
\label{fig:supp_text_cond_image_ldm}
\end{figure*}
\begin{figure*}[t]
    \centering
    \includegraphics[width=0.84\linewidth]{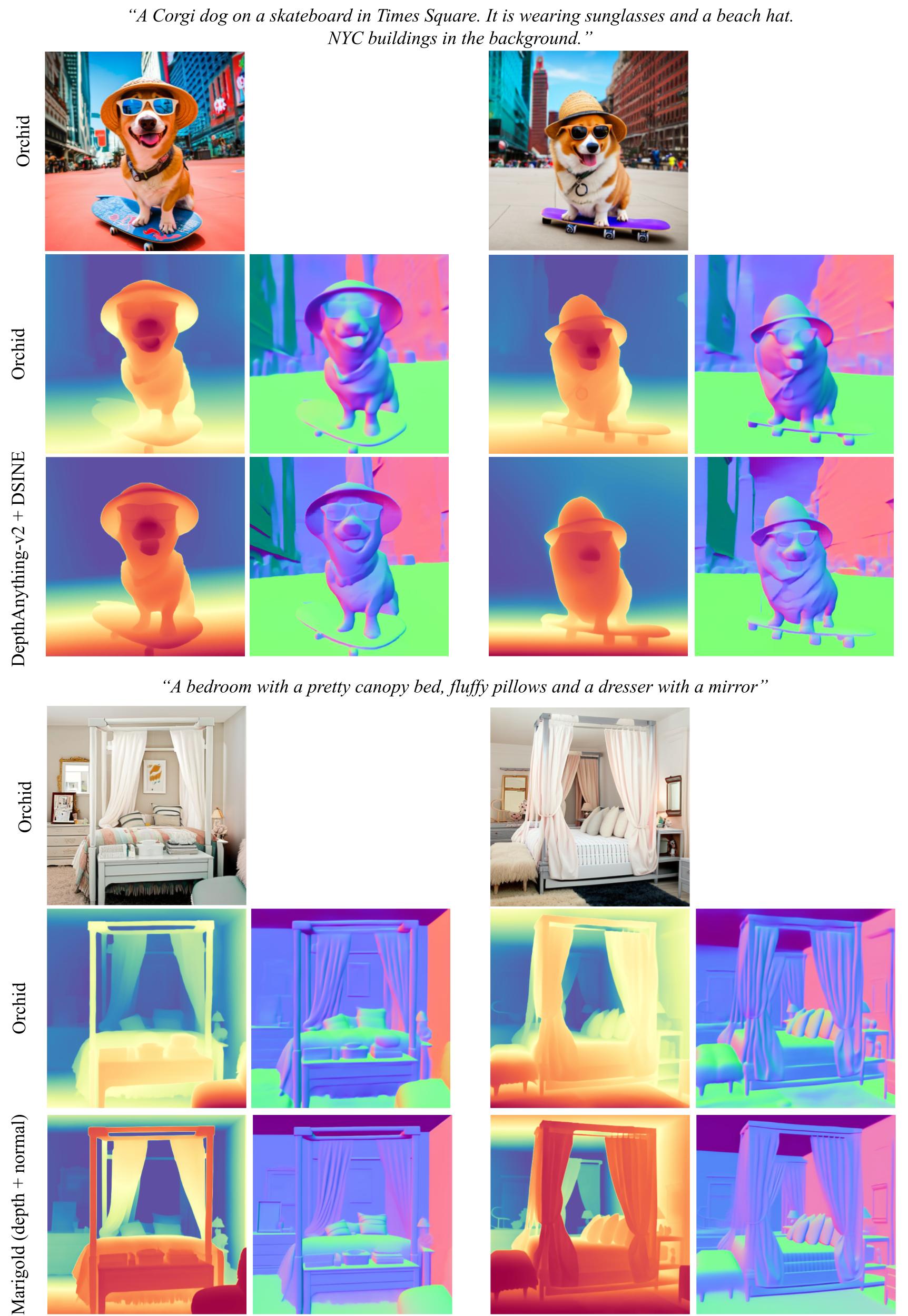}
    \cutcaptionup
    \caption
    {\textbf{Text conditioned color-depth-normal generation:} We show two predictions from \ourmodel{} for each text prompt. We compare the geometry predicted by our model to Marigold (separate) depth and normal models~\cite{marigold}, and to the DepthAnything-v2 + DSINE combination\cite{depth_anything_v2, bae2024dsine}. We find Orchid's geometry predictions to be qualitatively better, especially on structures / people in the background in the Corgi image. Color-conditional models may be inaccurate in such cases where geometry is ambiguous. When comparing results, please refer to our note on depth map visualization (Section~\ref{sec:note_normalization}). 
    }
    \cutcaptiondown
\label{fig:supp_text_cond_depth_mg}
\end{figure*}
\begin{figure*}[t]
    \centering
    \includegraphics[width=0.95\linewidth]{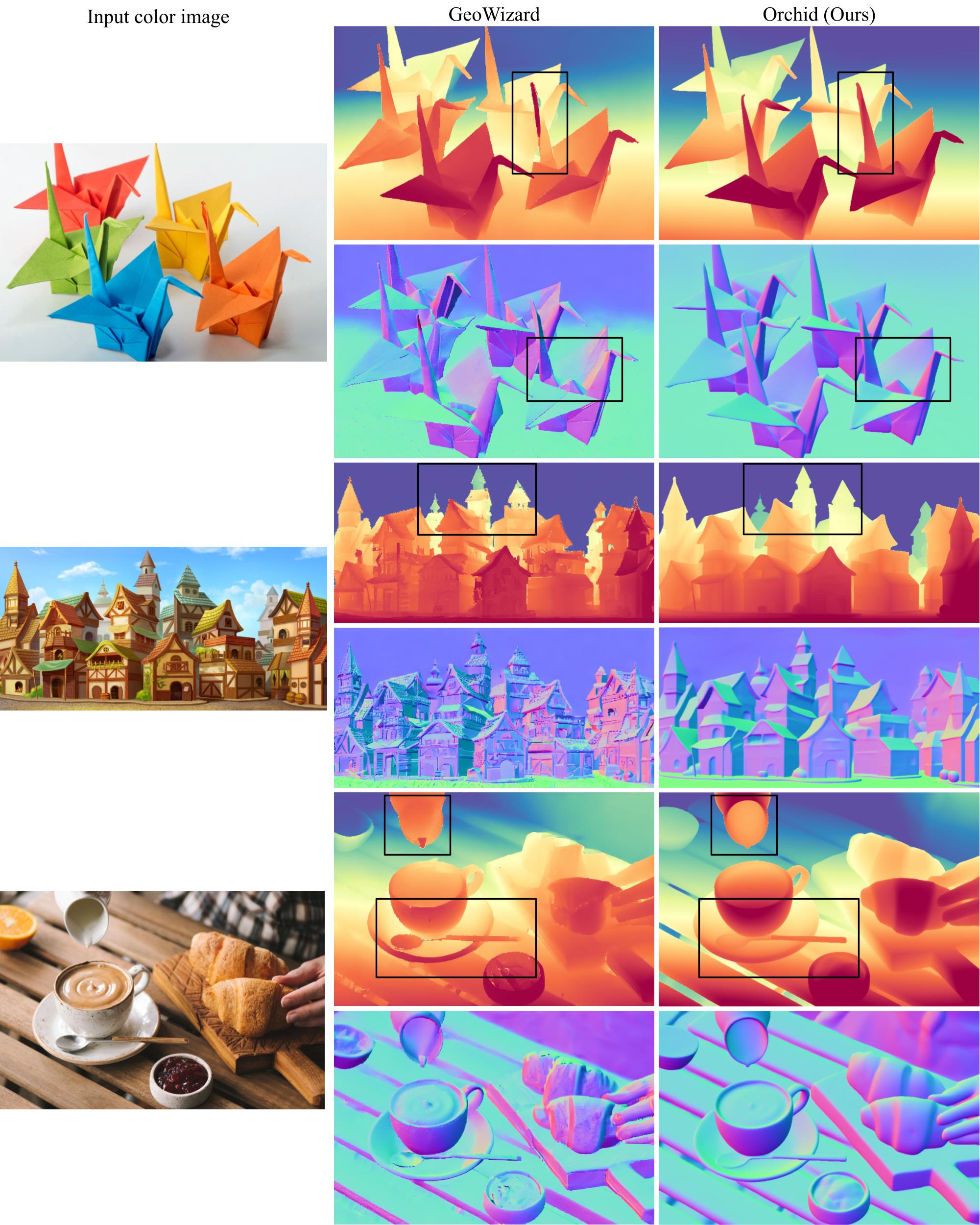}
    % \cutcaptionup
    \cutcaptionup
    \caption
    {Comparison of GeoWizard~\cite{fu2024geowizard} and \ourmodel{} for depth and normal estimation on in-the-wild input images. We can see that unlike GeoWizard, results from \ourmodel{} have correct depth and normal predictions while still having sharp boundaries. Some of these areas have been highlighted in the images shown above. In particular, Orchid shows less discontinuities in the Origami surfaces in both depth and normals, and more accurate depth predictions of the hollow objects pictured (milk pitcher, coffee mug and saucer).
    }
    % \cutcaptiondown
\label{fig:image_cond_ours_vs_geowizard_group1}
\end{figure*}

\begin{figure*}[t]
    \centering
    \includegraphics[width=0.90\linewidth]{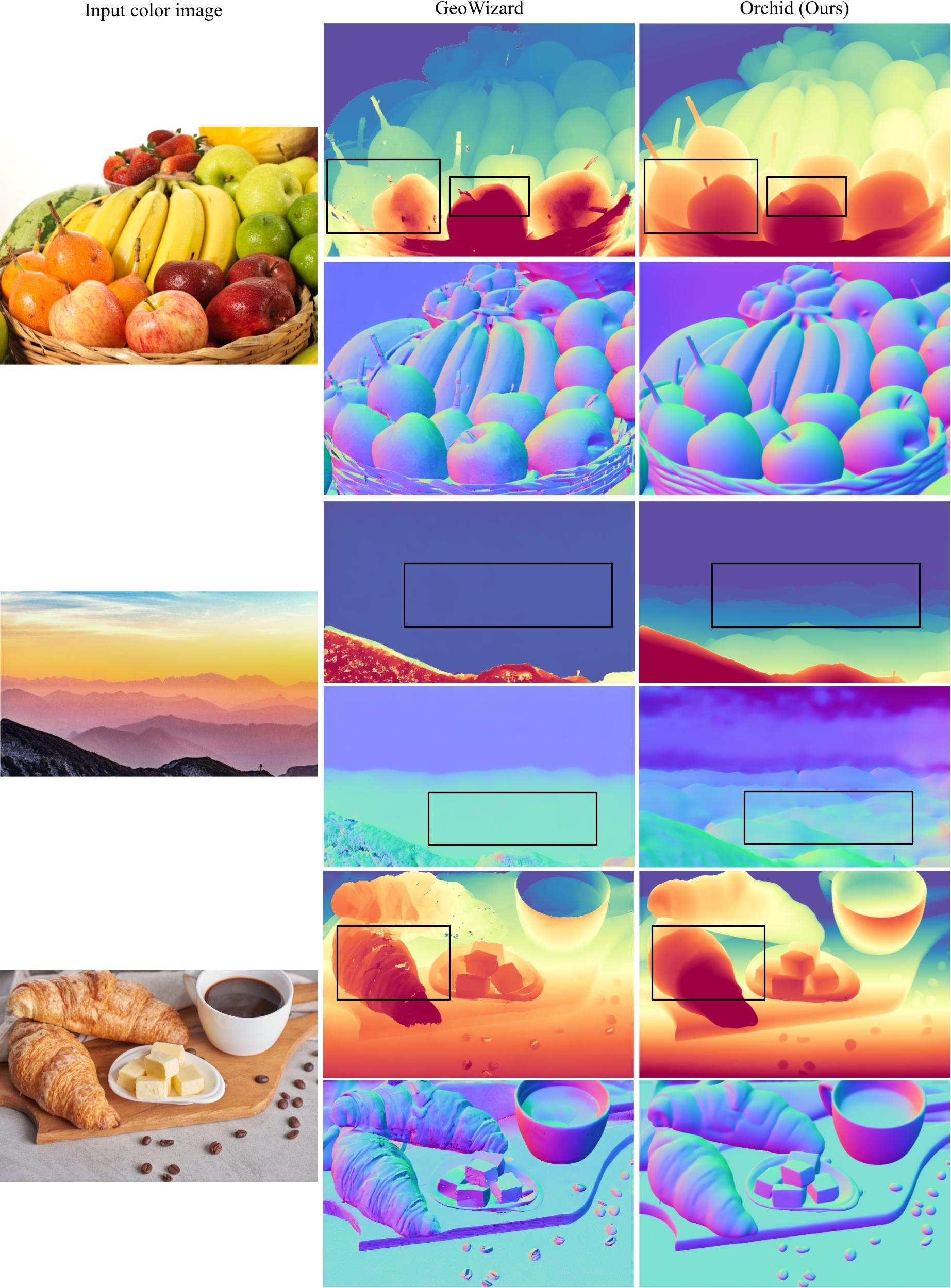}
    % \cutcaptionup
    \cutcaptionup
    \caption{Comparison of GeoWizard~\cite{fu2024geowizard} and \ourmodel{} on in-the-wild input images. Some areas with larger differences have been highlighted. In particular, we observe that high-frequency parts of the image can manifest themselves in noisy depth and normal predictions by GeoWizard (highlights on the fruits, texture of the croissants), whereas \ourmodel{} correctly predicts smooth surfaces. In far-away layered scenes we also observe that GeoWizard's predictions do not cover background (mountain range example).
    }
    % \cutcaptiondown
\label{fig:image_cond_ours_vs_geowizard_group2}
\end{figure*}

\begin{figure*}[t]
    \centering
    \includegraphics[width=0.94\linewidth]{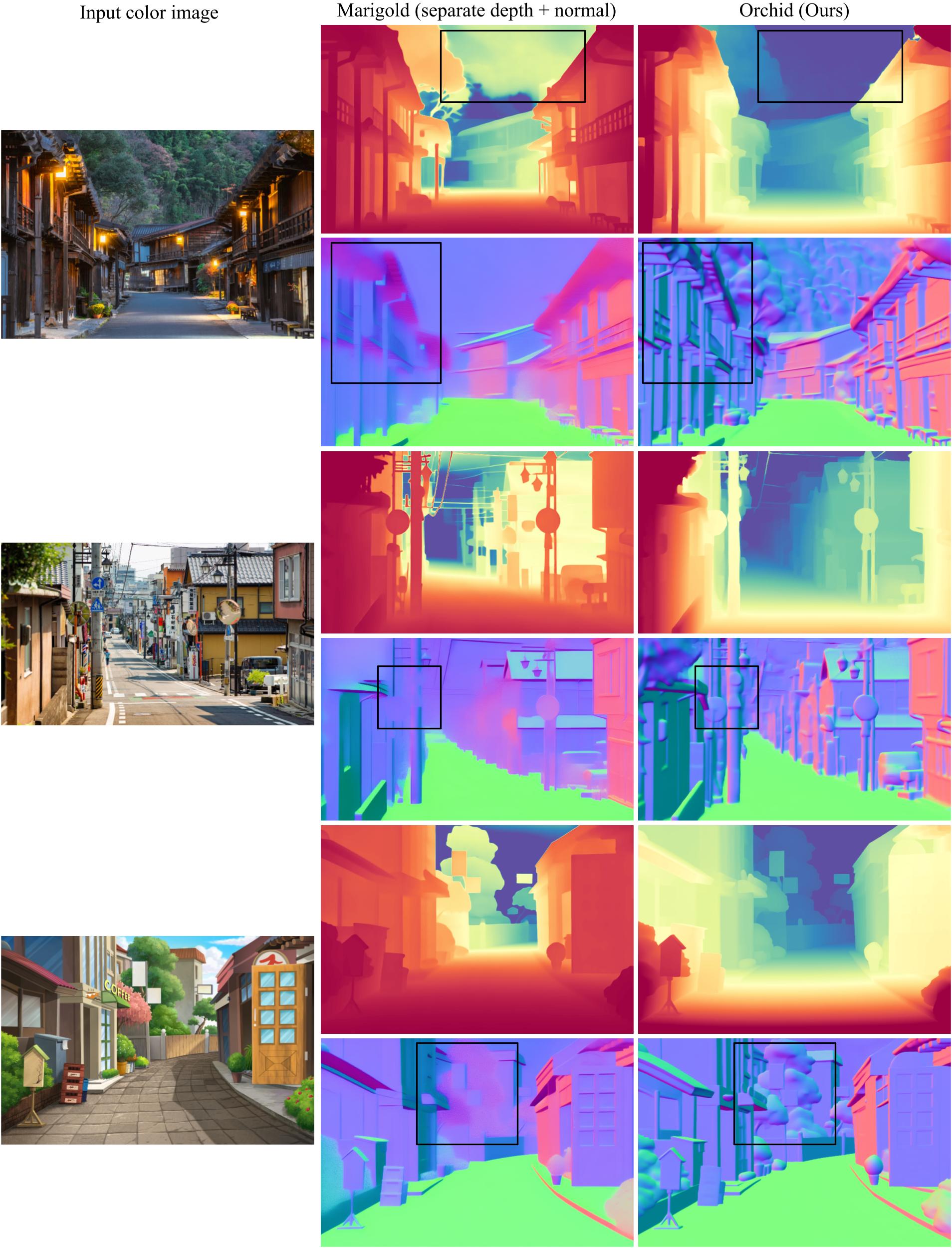}
    % \cutcaptionup
    \cutcaptionup
    % \vspace{-0.5em}
    \caption
    {Comparison of Marigold~\cite{marigold} and \ourmodel{} on some in-the-wild input images. We use separate Marigold models to predict depth and normals. \ourmodel{}'s joint predictions are better, especially for surface normals. Some notable differences are highlighted above.
    }
    % \cutcaptiondown
\label{fig:image_cond_ours_vs_marigold_group1}
\end{figure*}

\begin{figure*}[t]
    \centering
    \includegraphics[width=\linewidth]{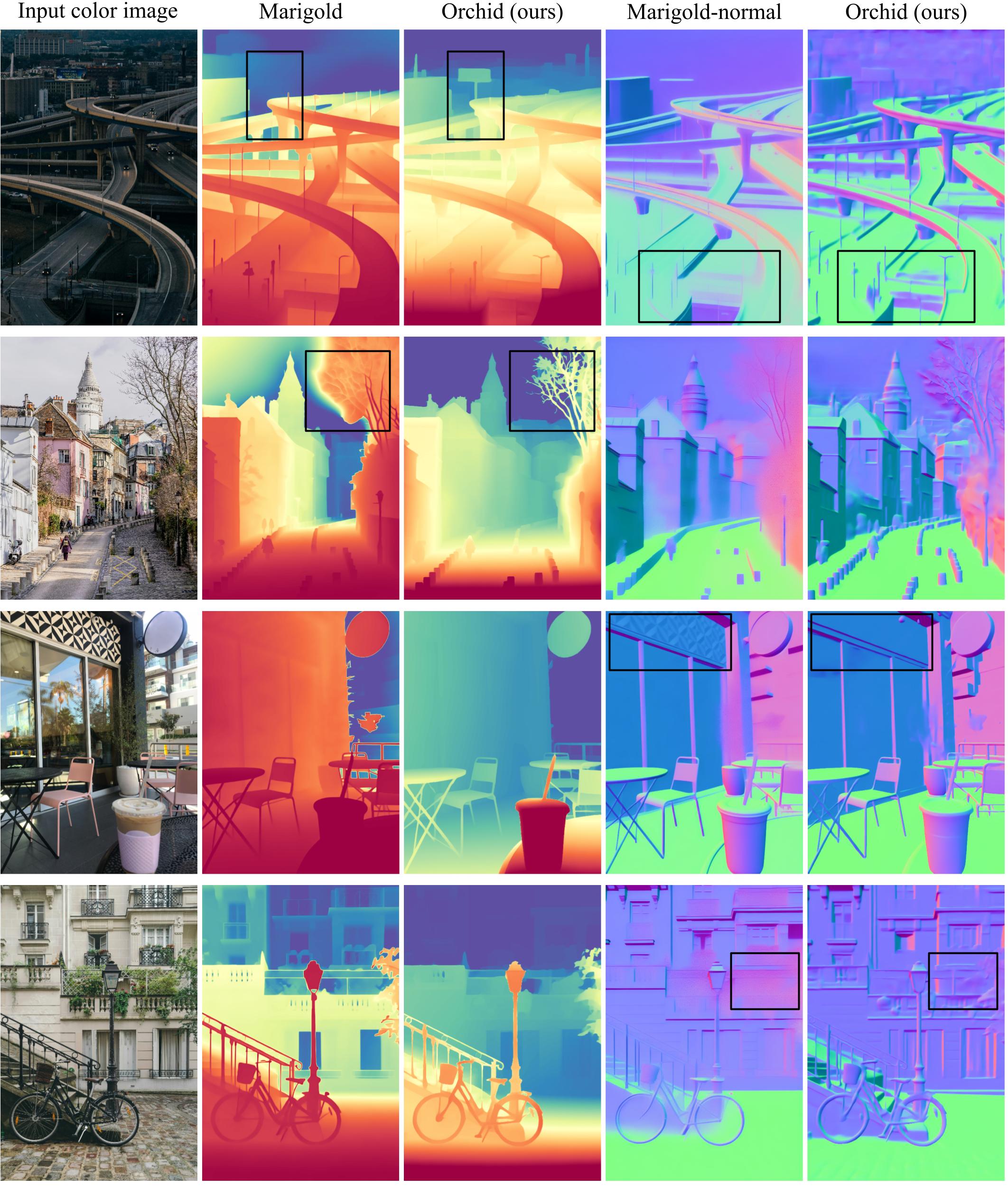}
    % \cutcaptionup
    % \cutcaptionup
    \caption
    {Comparison of Marigold~\cite{marigold} and \ourmodel{} on some in-the-wild input images. We can clearly see that our model \ourmodel{} can correctly predicts depth and surface normal of both far-away and nearby objects. Depth-maps from \ourmodel{} also has sharper and more accurate boundaries near pixels with depth discontinuities (\eg between narrow tree branches and sky). Some of these are highlighted in the figure above.
    }
    % \cutcaptiondown
\label{fig:image_cond_ours_vs_marigold_group2}
\end{figure*}

\begin{figure*}[h]
    \centering
    \includegraphics[width=0.95\linewidth]{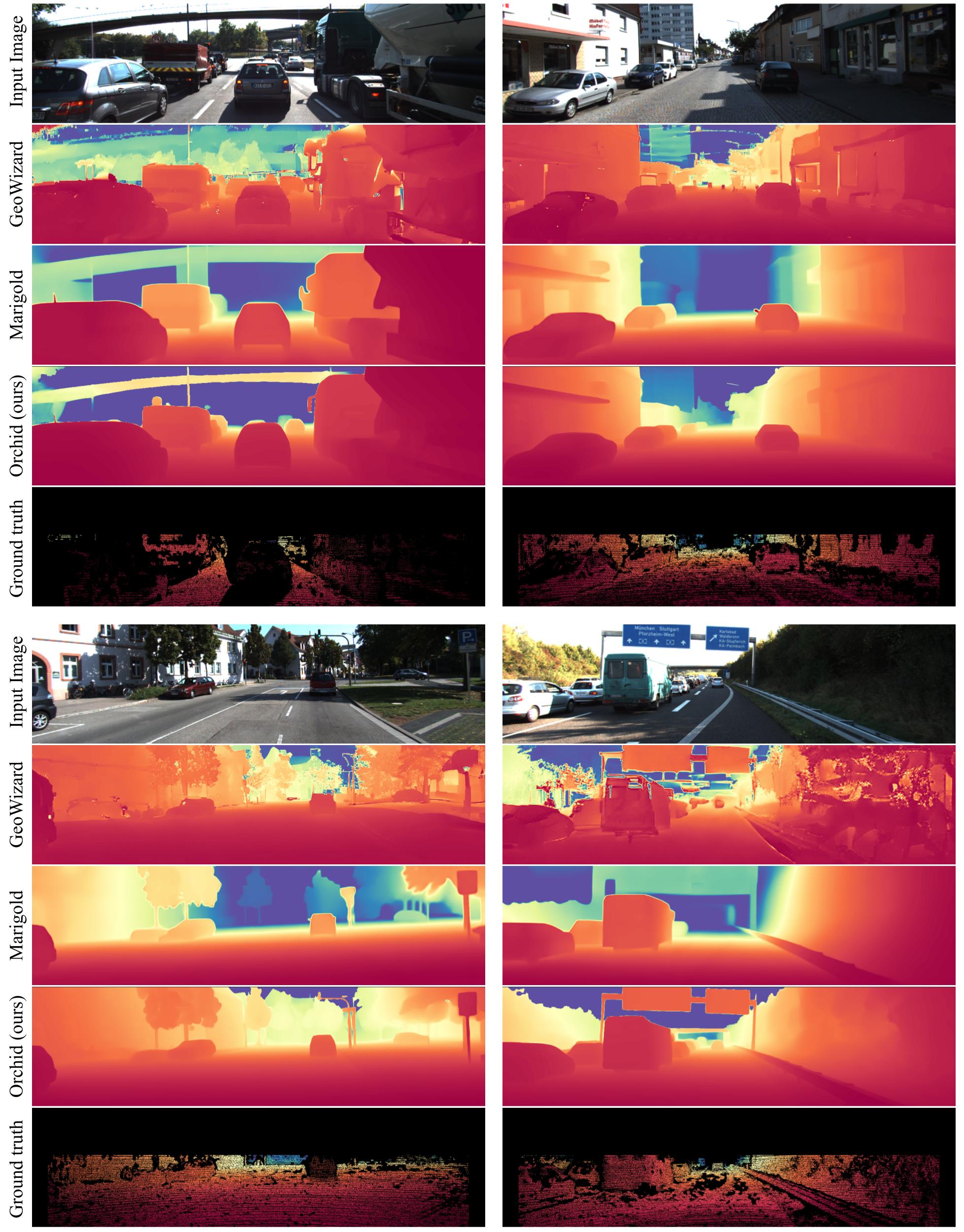}
    % \cutcaptionup
    \cutcaptionup
    \caption
    {Qualitative comparison of monocular depth prediction on KITTI~\cite{Geiger2012CVPR} dataset between GeoWizard~\cite{fu2024geowizard}, Marigold~\cite{marigold} and \ourmodel{}. Ground-truth depth (from lidar) are shown in the bottom row. Pixels without valid ground-truth depth are colored black. \ourmodel{}'s predictions are significantly better, especially at longer ranges.
    }
    \cutcaptiondown
\label{fig:mono_depth_kitti}
\end{figure*}

\begin{figure*}[t]
    \centering
    \includegraphics[width=0.98\linewidth]{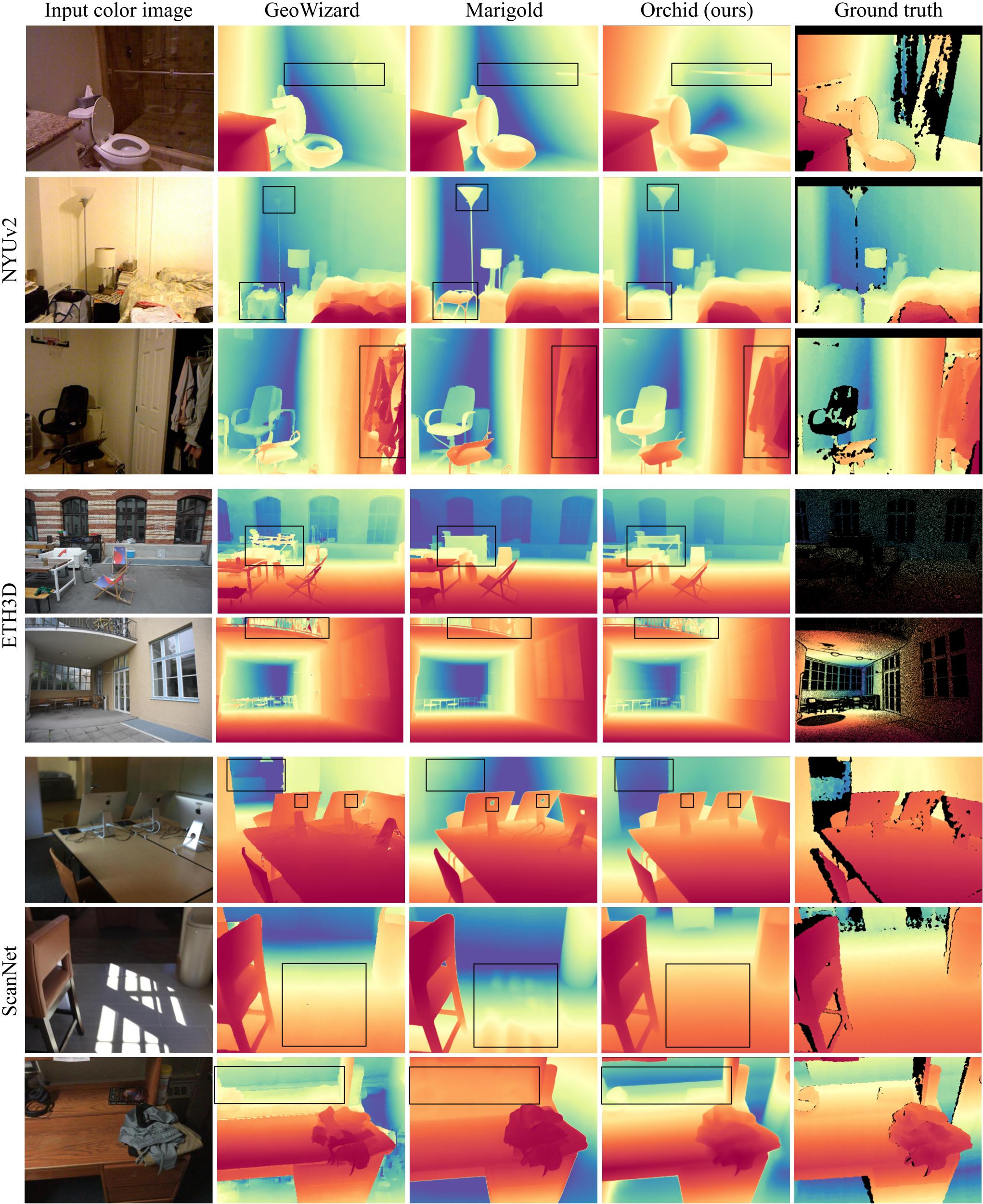}
    \cutcaptionup
    % \cutcaptionup
    \caption
    {Comparison of monocular depth prediction results by GeoWizard~\cite{fu2024geowizard}, Marigold~\cite{marigold} and \ourmodel{} on NYUv2~\cite{Silberman:ECCV12}, ETHD3D~\cite{schops2017multi}, and ScanNet~\cite{dai2017scannet} datasets. Ground-truth depth are shown in the rightmost column. Pixels without valid ground-truth depth are colored black. Our model \ourmodel{} has better depth predictions. Some notable differences are highlighted.
    }
    \cutcaptiondown
\label{fig:mono_depth_bench}
\end{figure*}

\begin{figure*}[t]
    \centering
    \includegraphics[width=\linewidth]{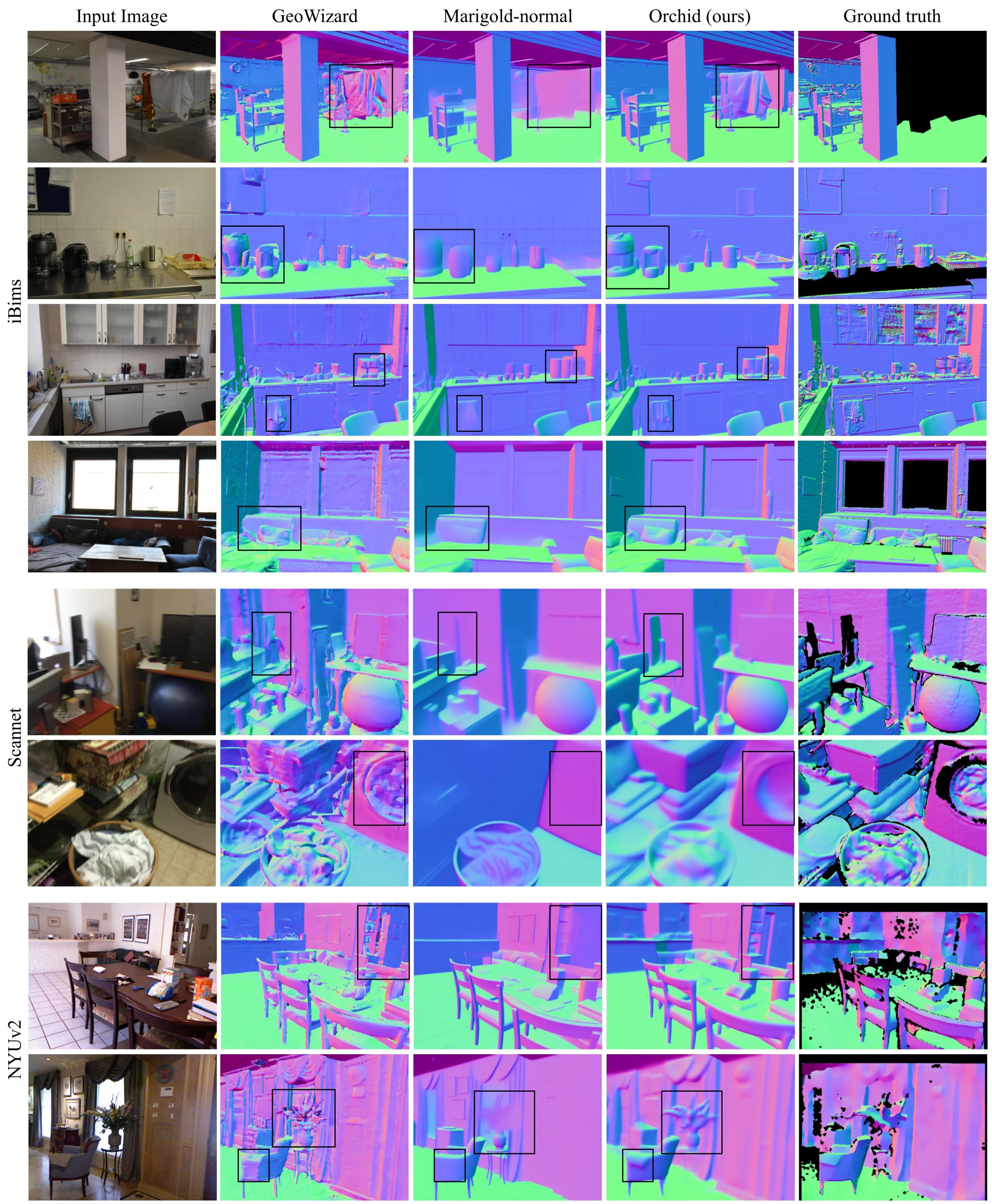}
    \cutcaptionup
    \cutcaptionup
    \caption
    {We compare single color image to surface-normal prediction methods of GeoWizard~\cite{fu2024geowizard}, Marigold~\cite{marigold} and \ourmodel{} on  iBims~\cite{koch2018evaluation}, and ScanNet~\cite{dai2017scannet}, and NYUv2~\cite{Silberman:ECCV12} datasets. Ground-truth normal are shown in the rightmost column. Pixels without valid ground-truth normal are colored black. Some notable differences are highlighted. \ourmodel{}'s normals are significantly better than baselines.
    }
    \cutcaptiondown
\label{fig:mono_normal_bench}
\end{figure*}
\begin{figure*}[t]
    \centering
    \includegraphics[width=0.8\linewidth]{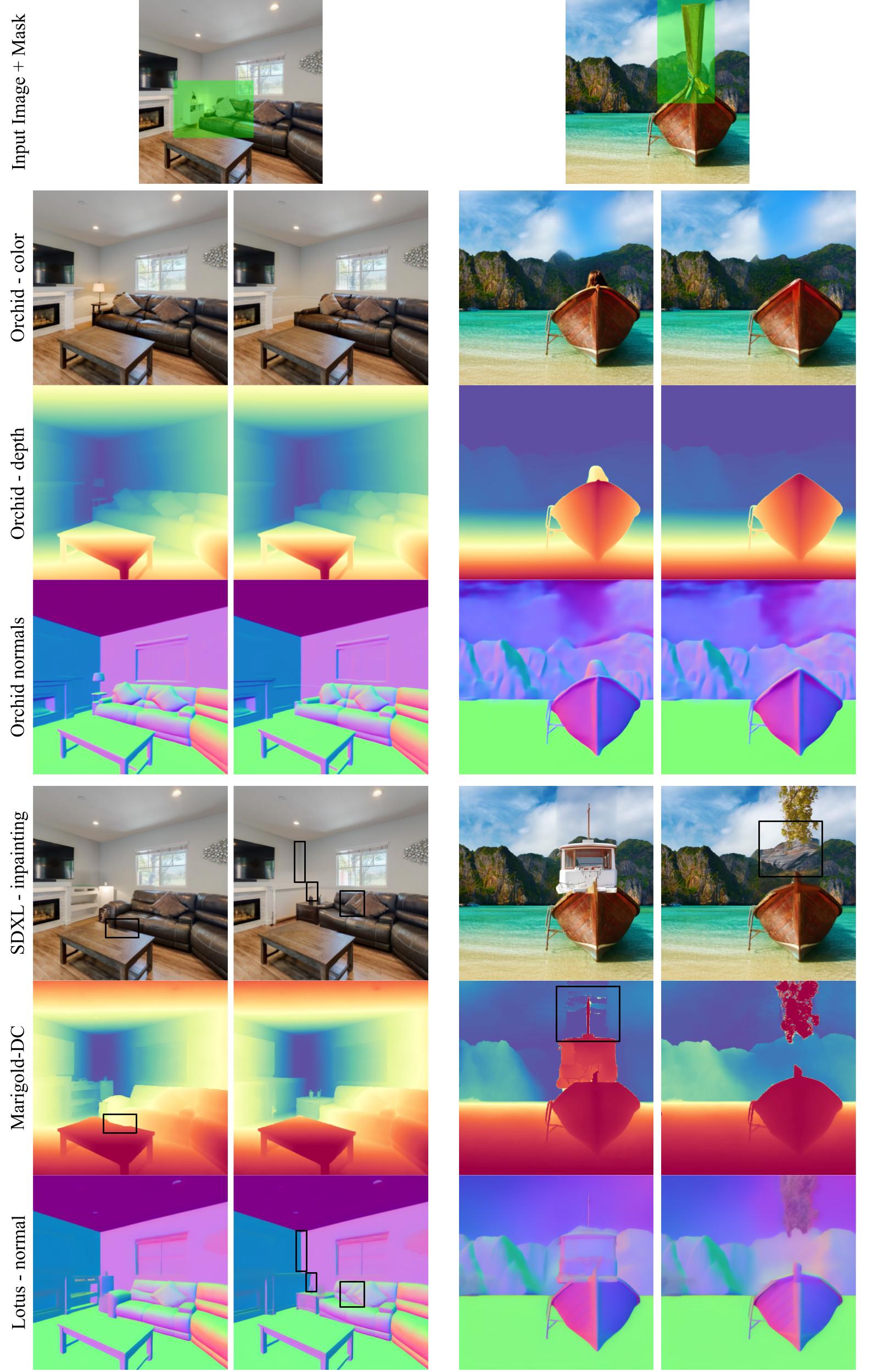}
    \cutcaptionup
    % \cutcaptionup
    \caption
    {\textbf{Joint color-depth-normal inpainting}: Given color-depth-normal images with masked regions, our model inpaints them jointly. Masked-out pixels are shown with green overlays on the input images. Inpainted outputs from \ourmodel{} look very realistic. For \eg, the edge of the wall is a continuous straight line, unlike the inpainting generated by a color-inpainting SDXL model. The inpainted results are also diverse (\eg the table lamp, the shape of the canoe).
    }
    \cutcaptiondown
\label{fig:supp_joint_inpainting}
\end{figure*}
% For \eg in top row, the inpainted pillow seems to also have a similar color theme as the other pillow and the sofa.

\begin{figure*}[t]
    \centering
    \includegraphics[width=0.9\linewidth]{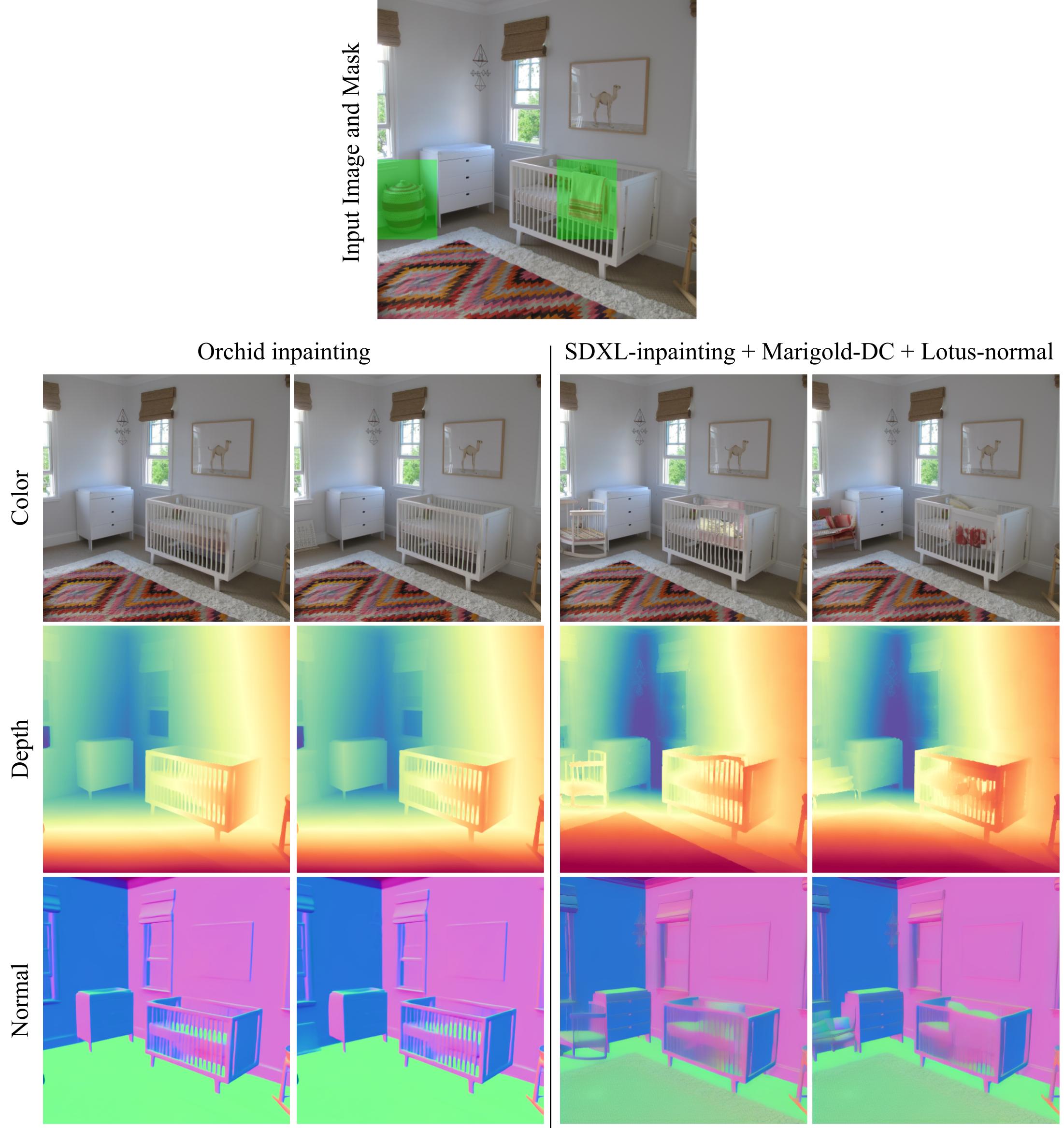}
    \cutcaptionup
    % \cutcaptionup
    \caption
    {\textbf{Joint color-depth-normal inpainting}: (contd.\ from Figure \ref{fig:supp_joint_inpainting}) Our model inpaints color-depth-normals them jointly. Masked-out pixels are shown with green overlays on the input image. Inpainted outputs from \ourmodel{} are much more realistic, including geometric details such as the shape of the cradle. On the other hand, multimodal inpainting using existing baselines produce geometric artifacts and unrealistic results. When comparing results, please refer to our note on depth map visualization (Section~\ref{sec:note_normalization}). 
    }
    \cutcaptiondown
\label{fig:supp_joint_inpainting_2}
\end{figure*}

\section{Color generation quality}

\begin{table}[h!]
\centering
\vspace{-1em}
\resizebox{\linewidth}{!}{
% BRISQUE score - 18.95, 20.62
\begin{tabular}{c | c c | c c}
\hline
\multirow{2}{*}{Method} & \multicolumn{2}{c|}{RGB generation metrics} & \multicolumn{2}{c}{User preference (\%)} \\
                        & CLIP ($\uparrow$)        & LPIPS ($\uparrow$)         & Aesthetics        & Text adherence  \\ \hline
RGB LDM (ours-PT)       & \textbf{0.319}   & 0.741          & 32.1  &  31.0                       \\
Orchid              & 0.316            & \textbf{0.764} & \textbf{47.4}  &  \textbf{46.9}                        \\ \hline
% SD-1.5    &         &       &      \\ \hline
No notable difference & - & -                           & 20.5  &  22.1  \\ \hline
\end{tabular}
}
\cutcaptionup
\caption{Quantitative evaluation and user study for RGB quality.}
\vspace{-1em}
\label{tab:supplm_rgb_quality}
\end{table}

While the focus of our work is not to improve the quality of generated color images, we evaluated how the quality of Orchid's text-conditioned color generations compare to the pretrained RGB-only diffusion model that we finetune from. In Table \ref{tab:supplm_rgb_quality}, we report the commonly used CLIP score and LPIPS for both models on the MS-COCO dataset, together with the findings of a user study we conducted. We generated images from both models using different captions, and asked users to pick from 3 options - Orchid's image, the base LDM's image, or notable difference. The users were asked to vote on two different aspects: aesthetics (overall quality of the image), text adherence (closeness to the text caption). We surveyed 40 users with 25 images each (1000 votes across both aspects in total). The quantitative metrics show that Orchid's generations are comparable to the color-only baseline, while the user study indicates that Orchid's generations are slightly better, with about 20\% votes indicating no notable difference between the two. These metrics depend significantly on the pretraining data and color-only model being used; Orchid maintains the pretrained generation quality while enabling joint color-depth-normal generation. 

\subsection{Note on depth-normal redundancy}

Using a joint latent for color, depth, and normals minimizes redundancy in our latent space, in comparison to using separate latents for each modality. Depth and normals are highly inter-dependent, as normals can be derived from (metric) depth. A joint latent avoids the need for separate latents, resulting in highly consistent predictions. To further validate this redundancy, we performed a PCA analysis on concatenated (separate) depth and normal latents (8 dimensions each, 1000 samples). Only 8 PCA bases (out of the full 16 dimensions) were needed to explain $>$ 95\% variance, confirming the strong depth-normal redundancy. 

\section{Qualitative results}
\label{sec:qualitative_results}

We provide additional qualitative results and comparisons for the experiments in our paper. Colormaps used to visualize the depth and surface normal predictions is shown in Figure \ref{fig:legend}. 

\subsection{Note on depth map visualization} 
\label{sec:note_normalization}
\cutsubsectiondown

\ourmodel{} predicts affine-invariant inverse depth, unlike other baselines Marigold \cite{marigold} and GeoWizard \cite{fu2024geowizard} that predict affine invariant depth normalized to  [0, 1]. To compare our depth qualitatively when ground truth depth is available (Figures \ref{fig:image_cond_ours_vs_geowizard_group1}, \ref{fig:image_cond_ours_vs_geowizard_group2}, \ref{fig:image_cond_ours_vs_marigold_group1}, \ref{fig:image_cond_ours_vs_marigold_group2}, \ref{fig:mono_depth_kitti}), we align all predictions to the ground truth by estimation a shift and scale offset using least squares. When ground truth is not available (Figures \ref{fig:supp_text_cond_image_ldm} and \ref{fig:supp_text_cond_depth_mg}), we inverted inverse-depth produced for our method, while using the predicted depth for \cite{marigold} and \cite{fu2024geowizard}, which may appear different due to an unknown inverse-depth shift. We use the colormap in Figure \ref{fig:legend}.

\subsection{Text conditioned joint generation}
\cutsubsectiondown

We show color-depth-normals generated by our model for different text prompts in Figure \ref{fig:supp_text_cond_ours}. Figure \ref{fig:supp_text_cond_image_ldm} compares the results from our model to a baseline that uses a color-only LDM to first generate color, and then depth and normal diffusion models to generate depth and surface normals. The results from a single pass of our model are comparable to these results. Figure \ref{fig:supp_text_cond_depth_mg} compares the depth and normals generated by our model to those predicted by depth and normal prediction baselines \cite{marigold, depth_anything_v2, bae2024dsine} on our images (generated along with depth and normals). In examples regions where depth is ambiguous for generated images (\eg background structure in Figures \ref{fig:supp_text_cond_image_ldm}, \ref{fig:supp_text_cond_depth_mg}), predictions from our model are qualitatively better. In other cases, our generated depth is comparable to those of baselines \cite{marigold} while our normals are significantly better.

\subsection{Monocular depth and normal estimation}
\cutsubsectiondown

\textbf{Internet images:} We show more depth and normal predictions on in-the-wild images produced by \ourmodel{} in Figures \ref{fig:image_cond_ours_vs_geowizard_group1}, \ref{fig:image_cond_ours_vs_geowizard_group2}, \ref{fig:image_cond_ours_vs_marigold_group1}, and \ref{fig:image_cond_ours_vs_marigold_group2}. Figures \ref{fig:image_cond_ours_vs_geowizard_group1} and \ref{fig:image_cond_ours_vs_geowizard_group2} compare our joint predictions to those from GeoWizard \cite{fu2024geowizard}. We find that our depth and normals are more accurate (with fewer errors on large sections), even though GeoWizard's predictions more detailed in many cases. In Figures \ref{fig:image_cond_ours_vs_marigold_group1}, and \ref{fig:image_cond_ours_vs_marigold_group2}, we compare \ourmodel{}'s predictions to Marigold \cite{marigold}. We find that \ourmodel{} has better depth estimates at longer ranges, and significantly better normal estimates overall. Note that we need different Marigold weights to predict depth and normals (unlike our joint prediction model). When comparing colorized depth maps on these datasets without ground truth depth, please refer to the note in Section \ref{sec:note_normalization}.

\noindent \textbf{Zero-shot benchmark images:} We show more depth and normal predictions on the zero-shot depth and normal estimation benchmarks used in Section 4 of our paper in Figures \ref{fig:mono_depth_bench}, \ref{fig:mono_depth_kitti}, and \ref{fig:mono_normal_bench}. Figure \ref{fig:mono_depth_bench} shows that  \ourmodel{} is competitive with diffusion-based depth prediction baselines Marigold \cite{marigold} and GeoWizard \cite{fu2024geowizard}, while being slightly better in some cases. Both \cite{marigold} and \cite{fu2024geowizard} have a common failure mode where depth estimates are sensitive to image discontinuities, which our model is significantly less sensitive to. Figure \ref{fig:mono_depth_kitti} shows that our model is significantly better are depth estimation in outdoor environments, especially at longer ranges. Figure \ref{fig:mono_normal_bench} shows that our model is significantly better at surface normals estimation, particularly on objects with curved surfaces. 

\subsection{Joint inpainting}\label{supp:joint_inpainting}
\cutsubsectiondown

Section 4.4 of our paper explains how our model can be used to jointly inpaint color-depth-normals. For this task, we use as input paired color, depth, and normal images, and a user-provided mask for the region to be inpainted. In cases where only a color image is available, depth and normals can be generated using the image-conditioned \ourmodel{}. We then generate the latents in the masked region, using \ourmodel{} to iteratively denoise them, while using noise-free latents encoded from the inputs for the unmasked region. This is similar to the approach proposed in RePaint  \cite{lugmayr2022repaint}.  We provide qualitative results in Figure \ref{fig:supp_joint_inpainting}. We show multiple inpainting results for the same input. We find that \ourmodel{} is able to generate very realistic images, with different semantically and geometrically consistent color, depth, and normals for the masked regions. We compare this to a baseline that first inpaints color (Stable-Diffusion XL-inpainting~\cite{podell2023sdxlimprovinglatentdiffusion}), then inpaints depth (Marigold-DC~\cite{viola2024marigolddc}), and predicts normals using Marigold/Lotus~\cite{marigold, he2024lotus}. The baseline performs significantly worse than \ourmodel{}, with several geometric inconsistencies in the generated color image (edges of objects or walls not intersecting, mismatch in vanishing directions, etc.). It also appears more unrealistic. The baseline uses conditional prediction on the full inpainted image for normals instead of inpainting them, as there are no publicly available normal-inpainting diffusion baselines. 

\subsection{3D reconstruction from single view}\label{supp:3d_recon}
\cutsubsectiondown

The image-conditioned \ourmodel{} can jointly generate depth and normals from an input image. These color, depth, and surface normals can be used to reconstruct the 3D scene using either Gaussian Splatting methods (3DGS~\cite{kerbl3Dgaussians}, 2DGS~\cite{Huang2DGS2024}) or Poisson surface reconstruction. The novel-view synthesis videos of reconstructions produced from the generated color and geometry are provided on our web page \textit{\href{https://orchid3d.github.io}{https://orchid3d.github.io}}. 

\section{Limitations and future work}

\ourmodel{} is not without limitations. In terms of geometry prediction accuracy, there is some scope for improvement on surfaces with high frequency edges (eg.: grass, fur, or hair). Some of these undesirably smooth predictions are apparent in our qualitative results on images in-the-wild. Future work can focus on further scaling unified appearance and geometry diffusion models, incorporating more recent developments in color diffusion models such as DiTs~ and flow-matching schedules. We also anticipate unified appearance-geometry diffusion models to be applied to many downstream reconstruction settings that are beyond the scope of our work: 3D scene completion, novel-view synthesis, text-conditioned full 3D generation, 3D inpainting, etc.

% --------------------------------------------------------------------------- %

\clearpage
{
    \small
    \bibliographystyle{ieeenat_fullname}
    \bibliography{main}
}

\end{document}